\documentclass{article}

\usepackage{microtype}
\usepackage{graphicx}
\usepackage{booktabs} 
\usepackage{enumitem}
\usepackage{subcaption}
\usepackage{microtype}
\usepackage{inconsolata}
\usepackage[T1]{fontenc}

\usepackage{hyperref}

\usepackage[accepted]{icml2026}

\usepackage{amsmath}
\usepackage{amssymb}
\usepackage{mathtools}
\usepackage{amsthm}
\usepackage{bbm}

\usepackage[capitalize,noabbrev]{cleveref}

\usepackage{fancyvrb}
\usepackage{multirow}
\usepackage{float}
\usepackage{listings}

\lstdefinestyle{prompt}{
  backgroundcolor=\color{gray!10},
  basicstyle=\ttfamily\small,
  basicstyle=\ttfamily\tiny,
  frame=single,
  rulecolor=\color{gray!60},
  breaklines=true,
  breakatwhitespace=false,
  columns=fullflexible,
  keepspaces=true,
  showstringspaces=false,
  tabsize=2,
}

\usepackage[capitalize,noabbrev]{cleveref}
\crefname{section}{\S}{\S\S}
\crefname{table}{Tab.}{Tab.}
\crefname{figure}{Fig.}{Figs.}
\crefname{algorithm}{Alg.}{}
\crefname{equation}{Eq.}{Eq.}
\crefname{appendix}{App.}{App.}
\crefname{theorem}{Theorem}{}
\crefname{restatableTheorem}{Theorem}{}
\crefname{prop}{Proposition}{}
\crefname{definition}{Def.}{}
\crefname{cor}{Corollary}{}
\crefname{observation}{Observation}{}
\crefname{assumption}{Assumption}{}
\crefname{hyp}{Hyp.}{Hypotheses}
\crefformat{section}{\S#2#1#3}
\crefname{namedtheorem}{Hyp.}{Hypotheses}

\newcommand{\states}{\mathcal{S}}
\newcommand{\actions}{\mathcal{A}}
\newcommand{\transitions}{\mathcal{T}}
\newcommand{\reward}{\mathcal{R}}
\newcommand{\obs}{\mathcal{O}}
\newcommand{\obsfunc}{\mathcal{Z}}
\newcommand{\policy}{\pi}

\newcommand{\lefta}{\textsc{Left}}
\newcommand{\righta}{\textsc{Right}}
\newcommand{\upa}{\textsc{Up}}
\newcommand{\downa}{\textsc{Down}}

\newcommand{\goalstate}{s_{\goal}}

\newcommand{\cell}[1]{\ensuremath{\mathsf{#1}}}
\newcommand{\wall}{\cell{wall}}
\newcommand{\agent}{\cell{agent}}
\newcommand{\open}{\cell{open}}
\newcommand{\goal}{\cell{goal}}
\newcommand{\padding}{\cell{padding}}

\usepackage[disable]{todonotes}

\definecolor{Blue}{RGB}{33,92,175}   
\definecolor{Green}{RGB}{98,115,19}      
\definecolor{PurpleDark}{RGB}{140,10,89} 
\definecolor{Purple}{RGB}{163,7,116} 
\definecolor{Gray}{RGB}{111,111,111} 
\definecolor{Red}{RGB}{183,53,45}   
\definecolor{Petrol}{RGB}{0,120,148}
\definecolor{Bronze}{RGB}{142,103,19} 
\definecolor{Orange}{RGB}{230, 100, 50}

\icmltitlerunning{A Behavioural and Representational Evaluation of Goal-Directedness in Language Model Agents}

\begin{document}

\twocolumn[
  \icmltitle{\centering A Behavioural and Representational Evaluation of\\Goal-Directedness in Language Model Agents}

  \icmlsetsymbol{equal}{*}

    \begin{icmlauthorlist}
        \icmlauthor{Raghu Arghal}{upenn,equal}
        \icmlauthor{Fade Chen}{nyu,equal}
        \icmlauthor{Niall Dalton}{ucl,equal}
        \icmlauthor{Evgenii Kortukov}{fraunhofer,equal}
        \icmlauthor{Calum McNamara}{indiana,equal}\\
        \icmlauthor{Angelos Nalmpantis}{tkh,equal}
        \icmlauthor{Moksh Nirvaan}{independent,equal}
        \icmlauthor{Gabriele Sarti}{northeastern}
        \icmlauthor{Mario Giulianelli}{ucl}
    \end{icmlauthorlist}
    
    \icmlaffiliation{upenn}{University of Pennsylvania}
    \icmlaffiliation{nyu}{New York University}
    \icmlaffiliation{fraunhofer}{Fraunhofer HHI}
    \icmlaffiliation{indiana}{Indiana University, Bloomington}
    \icmlaffiliation{tkh}{TKH AI}
    \icmlaffiliation{independent}{Independent}
    \icmlaffiliation{northeastern}{Northeastern University}
    \icmlaffiliation{ucl}{University College London}
    
    \icmlcorrespondingauthor{Mario Giulianelli}{\href{mailto:m.giulianelli@ucl.ac.uk}{m.giulianelli@ucl.ac.uk}}

  \icmlkeywords{goal-directedness, interpretability, evaluation, representation analysis, LLM agents, Machine Learning, ICML}

  \vskip 0.3in
]

\printAffiliationsAndNotice{$^*$Alphabetical order; see Statement of \nameref{sec:author-statement}.}  

\begin{abstract}
Understanding an agent's goals helps explain and predict its behaviour, yet there is no established methodology for reliably attributing goals to agentic systems. We propose a framework for evaluating goal-directedness that integrates behavioural evaluation with interpretability-based analyses of models' internal representations.
As a case study, we examine an LLM agent navigating a 2D grid world towards a goal state. 
Behaviourally, we evaluate the agent against optimal policies across varying grid sizes, obstacle densities, and goal structures, finding that performance scales with task difficulty while remaining robust to difficulty-preserving transformations and multi-goal structures.
We then use probing methods to decode internal representations of the environment and multi-step action plans. We find that the LLM agent non-linearly encodes a coarse spatial map, preserving approximate task-relevant cues about its position and the goal location; that its actions are broadly consistent with these internal representations; and that reasoning reorganises them, shifting from spatial cues towards immediate action selection.
Our findings support the view that introspective examination is required beyond behavioural evaluations to characterise how agents represent and pursue their objectives.\vspace{-1.1em}
\end{abstract}

\section{Introduction}\label{sec:intro}

\begin{figure}[!t]
    \centering
    \includegraphics[width=0.75\linewidth]{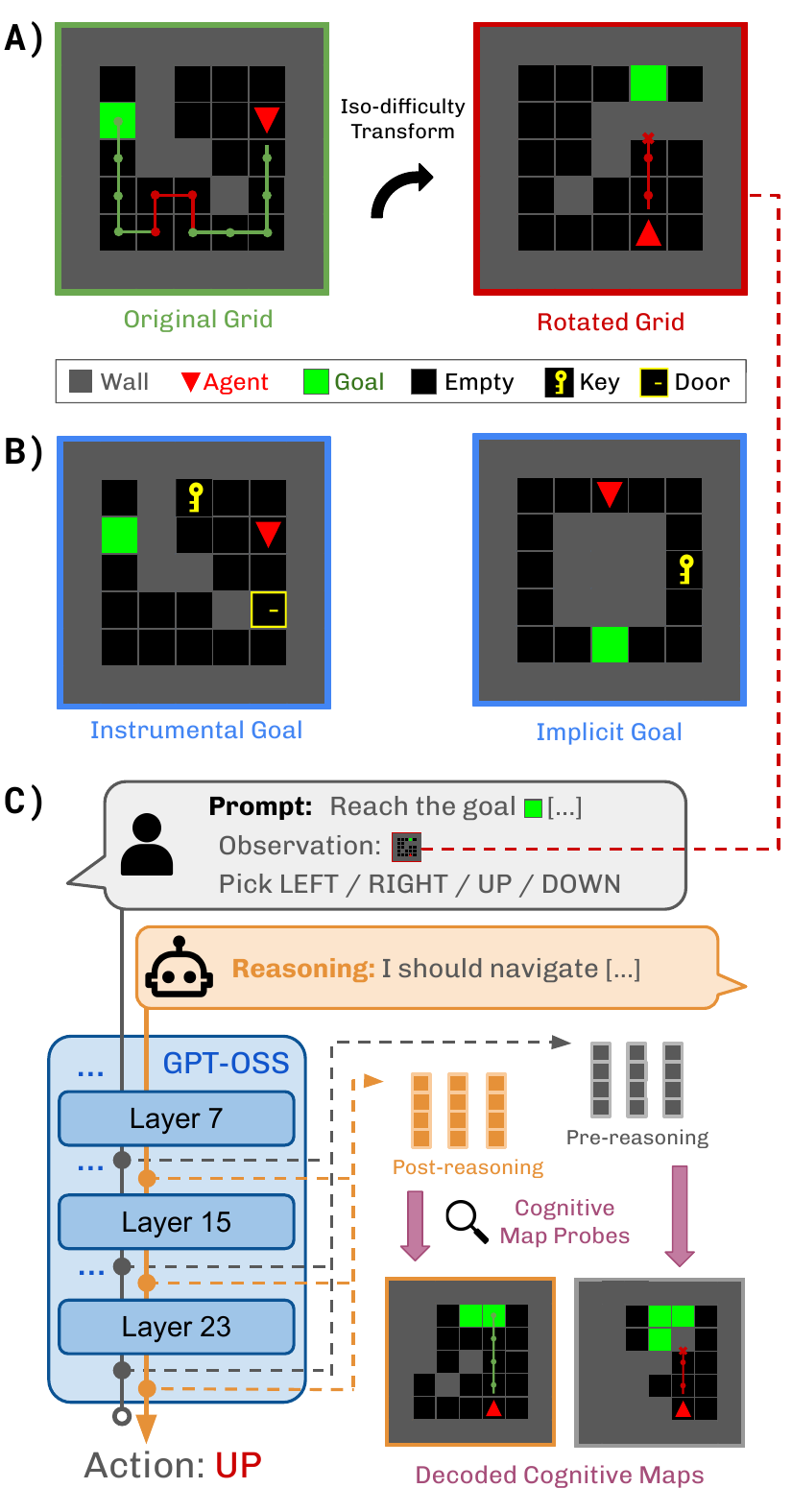}
    \caption{Overview of our goal-directedness analysis. We prompt an LLM-based agent to reason and act over a fully observable grid world. Behaviourally, we evaluate whether its trajectories \textcolor{Green}{agree} or \textcolor{Red}{disagree} with the optimal policy. \textbf{A:} Iso-difficulty transforms test whether the agent's behaviour is invariant to changes in grid configuration that preserve task difficulty. \textbf{B:} Multi-goal environments test how agents handle instrumental and implicit goal structures. \textbf{C:} We extract pre- and post-reasoning activations from intermediate layers, probe the agent's beliefs over goal distance, planned actions, and reconstruct \textit{cognitive maps} for the current grid state.}
    \label{fig:teaser}
    \vspace{-0.8em}
\end{figure}
Attributing goals to agents helps explain and predict their behaviour and provides a useful abstraction for reasoning about agency. 
Goal attribution has been studied across a wide range of fields, including philosophy \cite{davidson_radical_1973, dennett_interpretation_1990}, psychology and neuroscience \cite{baker2009, Schultz1997neural},
economics and decision theory \cite{vonneumann1944theory,savage_samuelsons_1948}, and reinforcement learning \cite{bellman1966dynamic,ng2000algorithms}. More recently, the question of when, and in what sense, goal attributions are warranted has become increasingly important for LLM-based agents \cite{xu-rivera-2024-measuring, macdermott_measuring_2024, everitt2025evaluating, goldstein_what_2025, mazeika_utility_2025}, particularly in the context of AI safety  \cite{naik2025misalignment, johnswentworth_instrumental_2025, marks2025auditing, li2025spilling, summerfield2025lessons,jarviniemi2026propensity}.

A natural way to measure goal-directedness is through \textit{behavioural evaluation}, i.e., assessing an agent's actions relative to a candidate goal, often by comparing them to an optimal policy \cite{xu-rivera-2024-measuring, everitt2025evaluating}. However, purely behavioural approaches face fundamental theoretical, practical, and philosophical challenges \cite{bellot2025limits, rajcicGoalDirectednessEyeBeholder2025, Chalmers_2025}.
First, agent capabilities may act as confounders for behavioural measures, as consistent failure may reflect capability limitations rather than an absence of goal-directed behaviour. Second, behavioural monitoring alone may be insufficient to guarantee alignment: a system with misaligned internal goals could produce aligned behaviour, or fail a safety-relevant evaluation task, when doing so is instrumentally useful \cite{hubinger2019risks, ngo2024the}.

To address these limitations, we propose a framework that combines behavioural evaluation with the analysis of internal representations, enabling holistic assessment of goal-directedness as a rich property arising from the interaction of beliefs, planning, and action selection. We study an LLM agent in a fully observable grid world, tasked with navigating to a goal state across grids of varying sizes and obstacle densities. 
We first subject the agent to standard capability tests, then introduce controlled environment perturbations and multi-goal task structures to assess the generalisability of its goal-directed behaviour. 
We find that the agent is sensitive to task difficulty and goal-like task-irrelevant cues, but robust to difficulty-preserving transformations and instrumental goals.
We then use activation probes to test if goal-relevant information can be decoded from the agent's internal activations before and after reasoning.
The probing analyses recover \textit{cognitive maps}---latent beliefs about the current environment state, including the agent position and the goal location---as well as planned multi-step action sequences.
We further find that task-relevant representations reorganise during reasoning: pre-reasoning activations preserve broader spatial cues and longer-horizon plans, while post-reasoning activations sharpen focus on next action selection. \Cref{fig:teaser} provides an overview of our approach.

\textbf{Contributions.}
Our primary contributions are as follows:\vspace{-10pt}\begin{enumerate}\itemsep0em
    \item We propose a white-box framework combining behavioural assessment and representation probing analyses for goal-directedness evaluation.
    \item We design controlled environment perturbations and multi-goal task structures to measure bias and robustness in the agent's goal-directed behaviour.
    \item We probe the agent's activations for environment state beliefs and multi-step action plans, and use these decoded representations to assess behavioural coherence relative to internal task-relevant information.
\end{enumerate}

Our code is available at \href{https://github.com/SPAR-Telos/interp}{github.com/SPAR-Telos/interp} and \href{https://github.com/SPAR-Telos/reveng}{github.com/SPAR-Telos/reveng}. We also provide an interactive viewer for decoded grids (\emph{cognitive maps}) at \href{https://huggingface.co/spaces/project-telos/trace-viewer}{huggingface.co/spaces/project-telos/trace-viewer}.

\section{Related Work}\label{sec:related_work}

The problem of identifying an agent's goals and intentions has a rich history spanning multiple research fields. 
Seminal work in philosophy \citep{davidson_radical_1973, lewis_radical_1974, dennett_interpretation_1990} and microeconomics \citep{vonneumann1944theory,savage_samuelsons_1948} has emphasised the predictive and explanatory power of assigning goals to an agent. 

\textbf{Measuring Agents' Goal-directedness.}
Recent work has sought to formally define and measure goal-directedness, particularly in the context of AI alignment and safety \citep{ward_reasons_2024,xu-rivera-2024-measuring,everitt2025evaluating,macdermott_measuring_2024}. 
Notably, \citet{everitt2025evaluating} define a measure of goal-directedness conditioned upon an agent's task-relevant capabilities and show goal-directedness is measurably distinct from performance in LLMs and generalises across tasks. \citet{macdermott_measuring_2024} build upon \citet{dennett_interpretation_1990}, proposing a formal measure of goal-directedness based on the predictive power of posited utility functions for the agent's behaviour. 
However, behavioural approaches to measuring goal-directedness are not without their weaknesses.
\citet{rajcicGoalDirectednessEyeBeholder2025} argue that such methods falter when faced with underspecification, coarse goals, uncertainty, and multi-agent settings. 
\citet{bellot2025limits} prove bounds on learnability from agent behaviour, showing that goal inferences are strictly limited by gaps between internal world models and the environment, as well as out-of-distribution shifts. Our work complements these approaches by enabling assessment of goal-directed behaviour relative to the agent's internal beliefs rather than ground truth alone. 

\textbf{Inverse Reinforcement Learning (IRL).} IRL is a direct instantiation of the goal attribution problem, aiming to infer a reward function from a policy or a set of demonstrations (see \citealp{ng2000algorithms,abbeel_apprenticeship_2004}, surveyed by \citealp{arora_survey_2021}). 
Work in this area has also been developed in connection with AI alignment (e.g., \citealp{hadfield-menell_cooperative_2016,hadfield-menell_inverse_2017}). 
While a weakness of classical IRL is the assumption that observed behaviour is optimal, approaches like MaxEnt IRL \citep{ziebartMaximumEntropyInverse2008a} aim to relax this via stochastic models of behaviour. 
Still, IRL methods suffer from the mis- and under-specification of the agent's behavioural model and latent reward function, respectively \citep{skalse_misspecification_2023}. Under strong assumptions---full observability, goal-directedness as optimal utility maximisation, perfect generalisation to new environments, and the observer's ability to perform interventions (i.e., to design environments and evaluate the agent within them)---behavioural experiments can, in principle, identify an agent's goal \citep{amin2016resolving}. However, these assumptions are unlikely to hold for LLM-based agents. In contrast to IRL, in this work we directly probe for goal-relevant representations without assuming a specific reward structure. 

\begin{figure*}[t!]
     \begin{subfigure}[b]{0.14\linewidth}
         \centering
         \includegraphics[width=\textwidth]{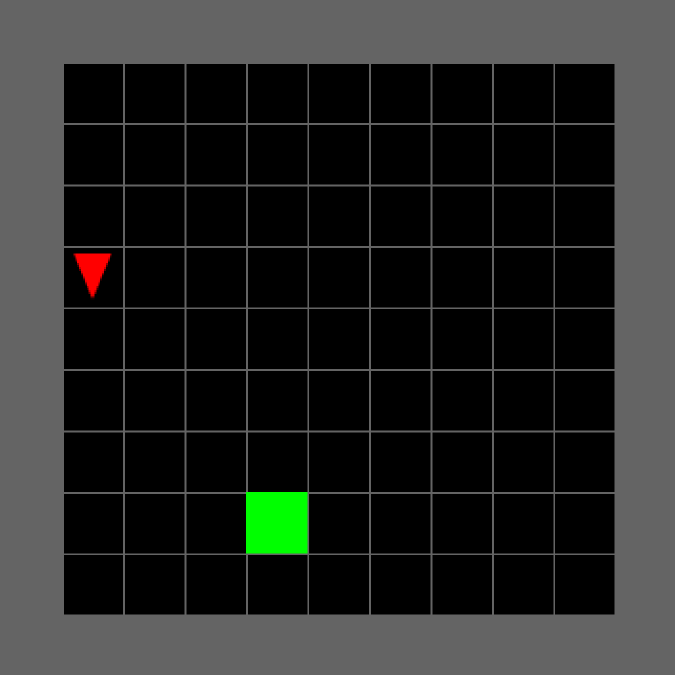}
         \caption*{$d=0.0$}
         \label{fig:first_grid}
     \end{subfigure}\hfill
     \begin{subfigure}[b]{0.14\linewidth}
         \centering
         \includegraphics[width=\textwidth]{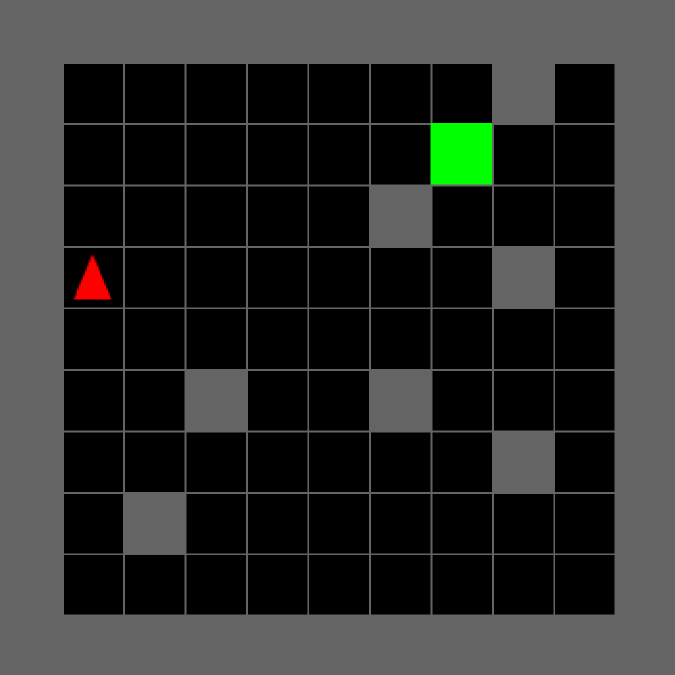}
         \caption*{$d=0.2$}
         \label{fig:second_grid}
     \end{subfigure}\hfill
     \begin{subfigure}[b]{0.14\linewidth}
         \centering
         \includegraphics[width=\textwidth]{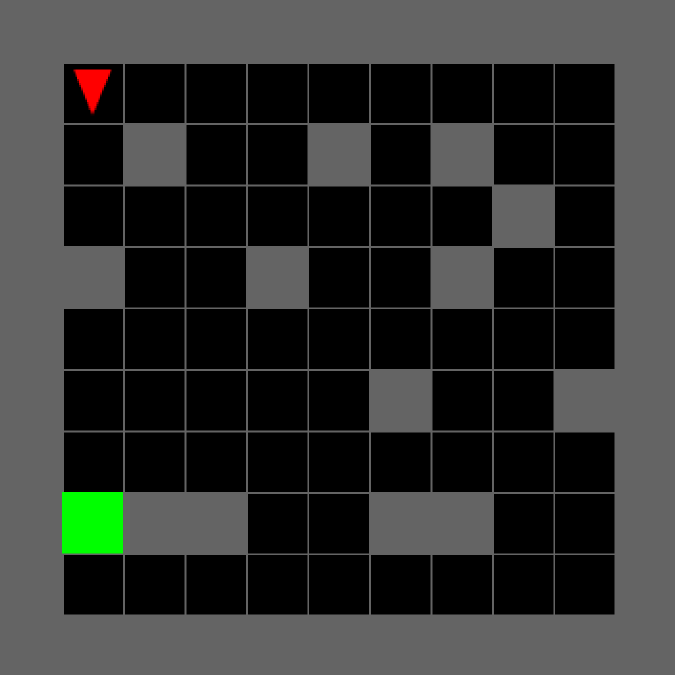}
         \caption*{$d=0.4$}
         \label{fig:third_grid}
     \end{subfigure}\hfill
          \begin{subfigure}[b]{0.14\linewidth}
         \centering
         \includegraphics[width=\textwidth]{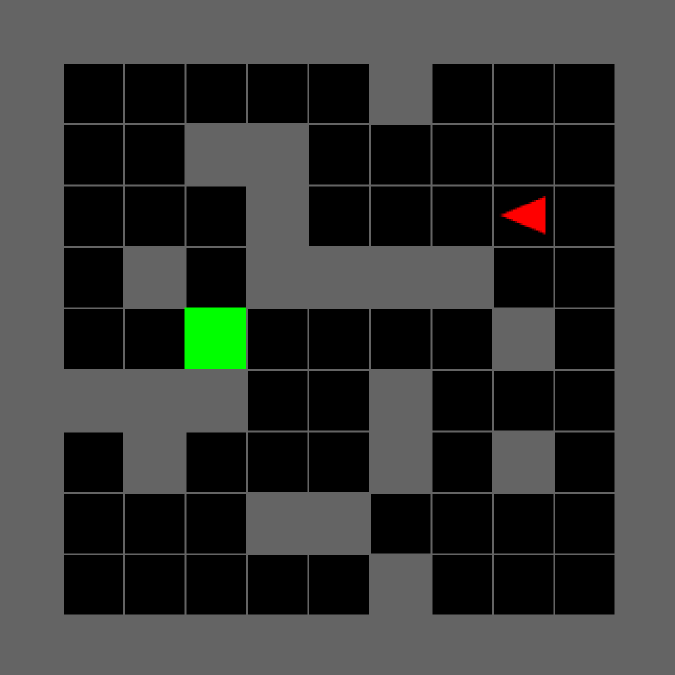}
         \caption*{$d=0.6$}
         \label{fig:fourth_grid}
     \end{subfigure}\hfill
          \begin{subfigure}[b]{0.14\linewidth}
         \centering
         \includegraphics[width=\textwidth]{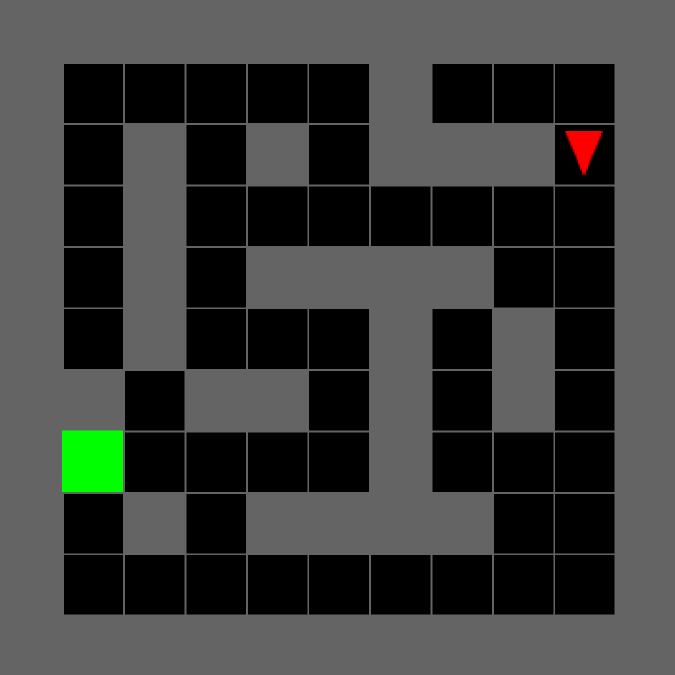}
         \caption*{$d=0.8$}
         \label{fig:fifth_grid}
     \end{subfigure}\hfill
          \begin{subfigure}[b]{0.14\linewidth}
         \centering
         \includegraphics[width=\textwidth]{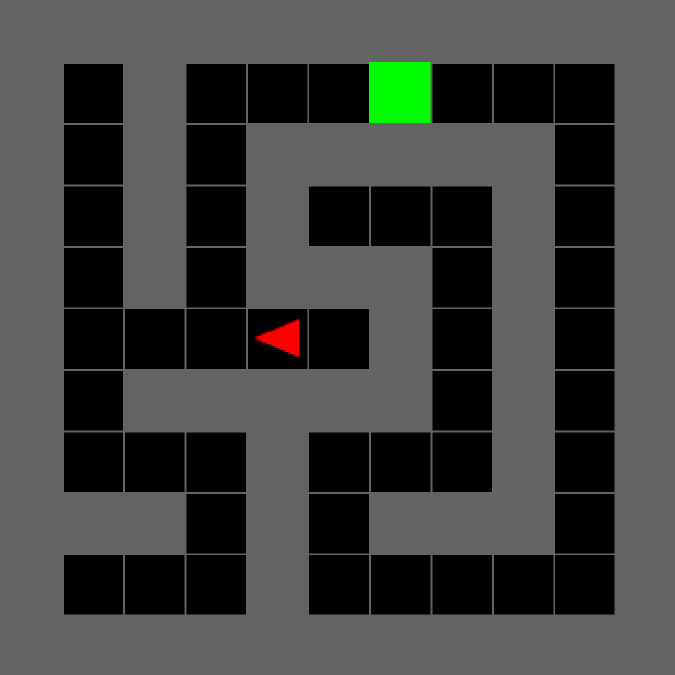}
         \caption*{$d=1.0$}
         \label{fig:sixth_grid}
     \end{subfigure}  
     \hfill 
     \caption{Grid worlds with increasing wall density $d$, from fully open grids ($d=0$) to maze-like grids with no circular paths ($d=1$).}
     \label{fig:complexities}
    \vspace{-0.8em}
\end{figure*}

\begin{figure}[t!]
     \centering
     \begin{subfigure}[b]{0.39\linewidth}
         \includegraphics[width=\textwidth]{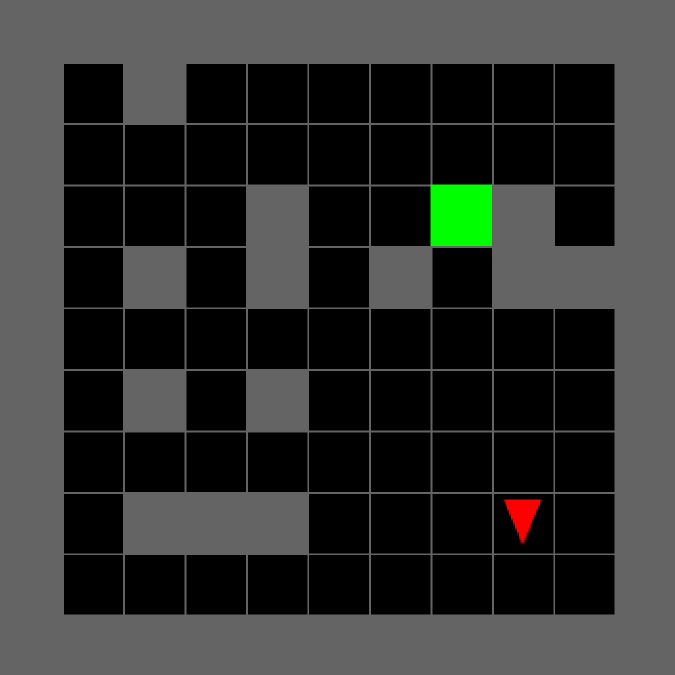}
         \label{fig:text_reps_grid}
     \end{subfigure}
     \hspace{1.5em}
     \begin{subfigure}[b]{0.39\linewidth}
         \includegraphics[width=\textwidth]{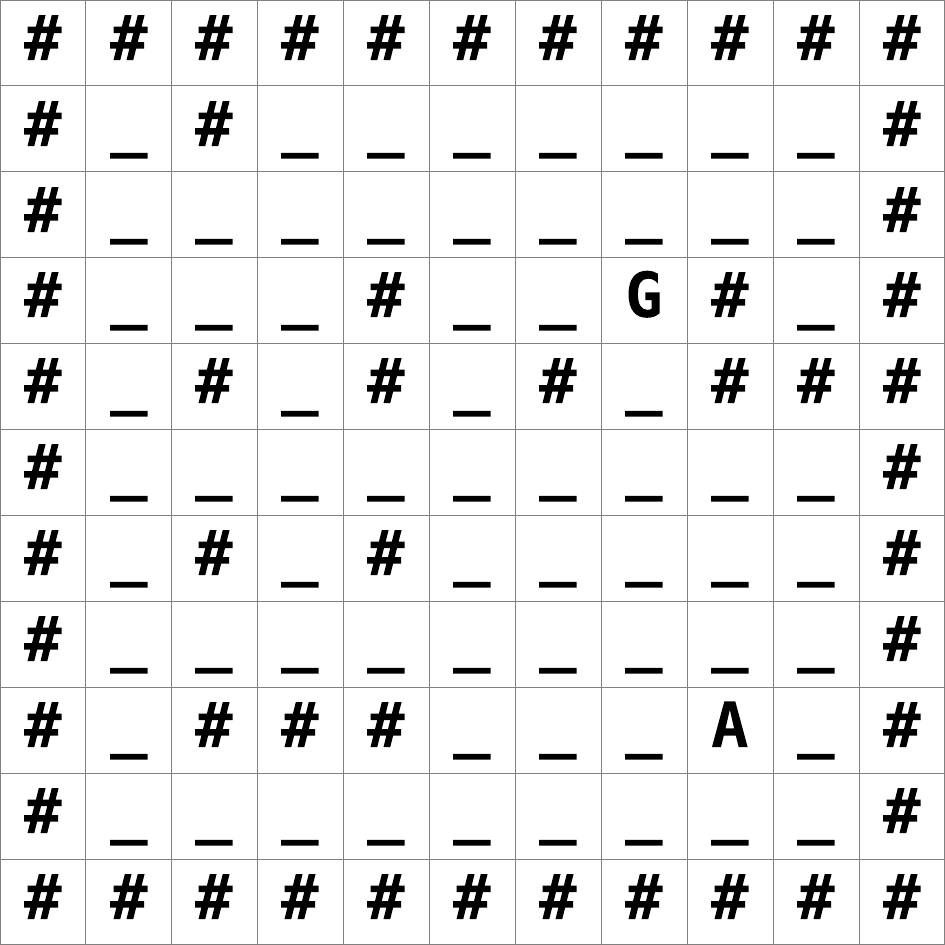}
         \label{fig:text_reps_text}
     \end{subfigure}
     \vspace{-0.8em}
     \caption{An example grid (left) and its corresponding text-based representation (right) used for LLM prompting.}
     \label{fig:text-representation}
     \vspace{-1.2em}
\end{figure}

\textbf{Probing Environment and Plans in LLMs.} Several studies have examined whether language models learn structured representations of their environment. \citet{li2023emergent} show that a GPT model trained to predict Othello moves develops a causally relevant representation of the board state, while \citet{nanda-etal-2023-emergent} show this representation can be decoded linearly. Similar linear representations of spatial and temporal information were found in LLMs trained on natural text \cite{gurnee2024language}. 
Recent work has also probed LLMs for goal-oriented abstractions \cite{li2024do} and shown that models engage in forward planning, pre-selecting future outputs before generating intermediate tokens \cite{pal-etal-2023-future,men-etal-2024-unlocking,lindsey2025biology,dong2025emergent}. 
Similar phenomena have also been observed in other neural architectures \cite{bush2025interpreting,taufeeque2025plannin}.
More broadly, high-level features have been found to be decodable from model activations and used for monitoring and steering \cite{li-etal-2021-implicit,zou2023representation,marks2024the}. We extend this line of work through the lens of propositional interpretability \cite{Chalmers_2025}, eliciting environment representations and plans from model internals in an agentic navigation task.

\section{Grid World Environment Setup}\label{sec:fo_gridworld}

We select GPT-OSS-20B \cite{openai2025gptoss120bgptoss20bmodel} for our evaluation in light of its manageable size and strong performance on complex tasks, and test it on a two-dimensional navigation task using the MiniGrid environment \cite{MinigridMiniworld23}. The LLM agent has full observability of the grid and is tasked with navigating to the goal square one action at a time. We translate the grid into a text-based representation (\Cref{fig:text-representation}) ensuring that each cell in the grid corresponds to exactly one token to limit issues arising from tokenisation \cite{edman-etal-2024-cute,cosma-etal-2025-strawberry}.

\textbf{Full Observability.} A fully observable environment offers a simple but tightly controlled setting for analysing goal-directedness, for at least three reasons.
(1) With full observability, the agent directly observes the true world state; this eliminates the need to maintain beliefs over hidden world states and allows optimal policies to be derived using standard algorithms.
(2) Full observability also removes several factors that might otherwise confound the analysis, including memory, belief updating under perceptual uncertainty, and exploration–exploitation trade-offs.
(3) We observed that LLM-based agents, including frontier models, perform poorly in partially observable grid worlds, exhibiting behaviours like redundant backtracking. This makes it difficult to disentangle capability limitations from failures of goal-directedness; we discuss this further in App.~\ref{app:po_appendix}.

\textbf{Grid Worlds.} We model $n\!\times\!n$ grid world environments as Markov Decision Processes defined by the tuple $(\states, \actions, \transitions, r, \gamma)$.
The state space is $\states = [n]^2$, representing grid locations, and the action space is $\actions = \{\upa, \downa, \lefta, \righta\}$.
Transitions are deterministic: the transition function $\transitions(s' \mid s, a) = \mathbb{P}(S_{t+1}=s' \mid S_t=s, A_t=a)$ moves the agent to the adjacent cell as determined by $a$ if that cell is open; otherwise (e.g., if the action would enter a wall), the agent remains in its current location.
A grid world instance is specified by a function
$
G : [n]^2 \rightarrow \{\wall, \open, \goal\}
$,
which assigns a cell type to each grid location. 
Grid worlds vary in obstacle density $d \in [0,1]$,  
where $d=0$ corresponds to a fully open grid and $d=1$ to a maze-like grid with no circular paths.
Examples of grids with different density levels are shown in \Cref{fig:complexities}.

\textbf{Policies and Trajectories.}
We write $\pi^*$ for an optimal policy, assumed to be uniform over optimal actions when multiple optima exist. An agent parameterised by $\theta$ follows a policy $\pi_{\theta}(a\mid s)=\mathbb{P}(A_t=a\mid S_t=s),$ with $a_t$ denoting the action selected by the agent at time $t$. Given a policy $\policy$ and an initial state $s_0$, a trajectory $\tau^{\policy}(s_0) = (s_i, a_i)_{i=0}^{T}$ is generated by executing $\policy$ from $s_0$. A trajectory is successful if the final state satisfies $s_T = \goalstate$; otherwise, it terminates upon reaching a fixed maximum horizon $T$.

\section{Behavioural Evaluation}
\label{sec:objective_policy_eval}

We begin with a behavioural evaluation of the agent's policy, comparing it against an optimal reference policy derived using $A^*$ search with Manhattan distance to the goal. This analysis assesses how closely the agent's action choices and action distributions align with optimal behaviour in the grid world, without inspecting the agent's internal representations.
We construct a set of grid worlds~$\mathcal{G}$ with sizes $\mathcal{Z}=\{7,9,11,13,15\}$ and obstacle densities $\mathcal{D}=\{0.0,0.2,0.4,0.6,0.8,1.0\}$. For each size–density pair in $\mathcal{Z} \times \mathcal{D}$, we generate 10 random grids. On each grid, the agent is evaluated over 10 trajectories using a sampling temperature of $0.7$ and a maximum horizon of $T=1.5 \times L$ where $L$ is the optimal path length to the goal.\footnote{We limit trajectory length to $1.5 \times L$ steps to filter cases where the agent is stuck moving back and forth between states.} 
Additional evaluation settings and prompts are reported in App.~\ref{app:settings}.

\subsection{Goal-Directedness across Baseline Task Conditions}
\label{subsec:baseline_tasks}

We report \emph{per-action accuracy}, defined as the fraction of actions along a trajectory that are optimal, i.e., $a_t \in \arg\max_{a \in \actions} \pi^*(a \mid s_t)$, the \textit{entropy} of the agent's action distributions, and the \textit{Jensen–Shannon divergence} (JSD) from the optimal policy. 
All metrics are averaged across trajectories and grids. 
To estimate the agent's policy $\pi_{\theta}$, we compute empirical action probabilities from the relative frequency of actions across all available trajectories for a given grid.\footnote{We use relative frequency instead of action token log-probability since the latter converges to 1 after the model reasoning chain, making it a poor proxy for agent uncertainty.}
App.~\ref{app:behavioural-eval-metrics} provides formal definitions and introduces additional metrics, including the success rate and the expected calibration error, with full results shown in App.~\ref{app:additional_objective_results}. While relevant, we do not find these to provide additional insight beyond those discussed here.

\begin{figure}[t!]
    \centering
    \includegraphics[width=\linewidth]{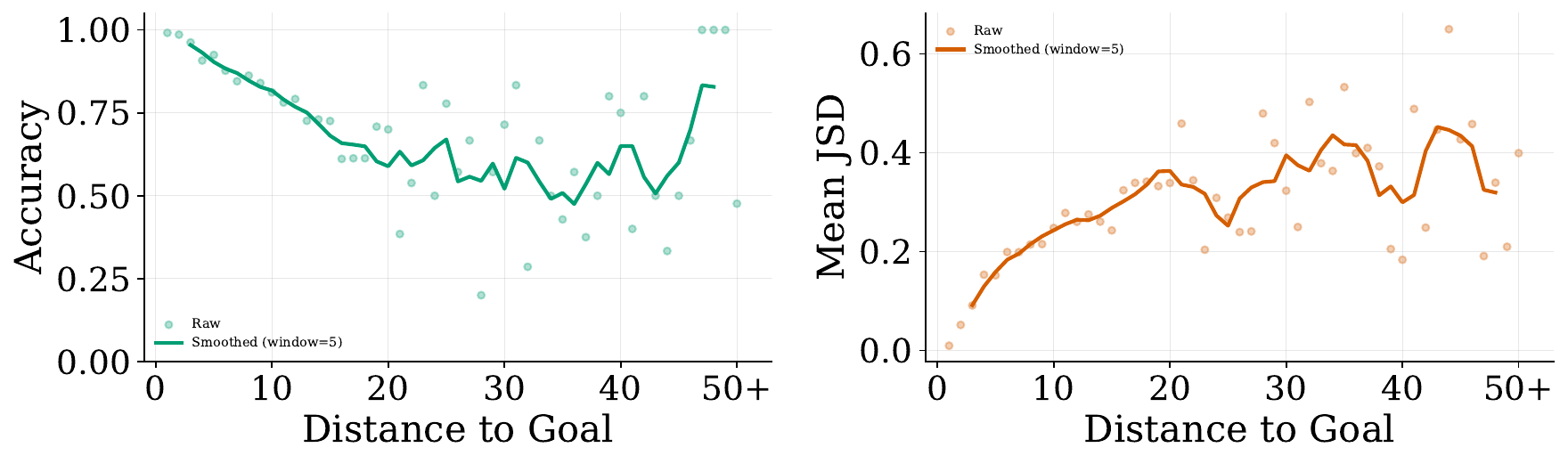}
    \hfill
    \includegraphics[width=\linewidth]{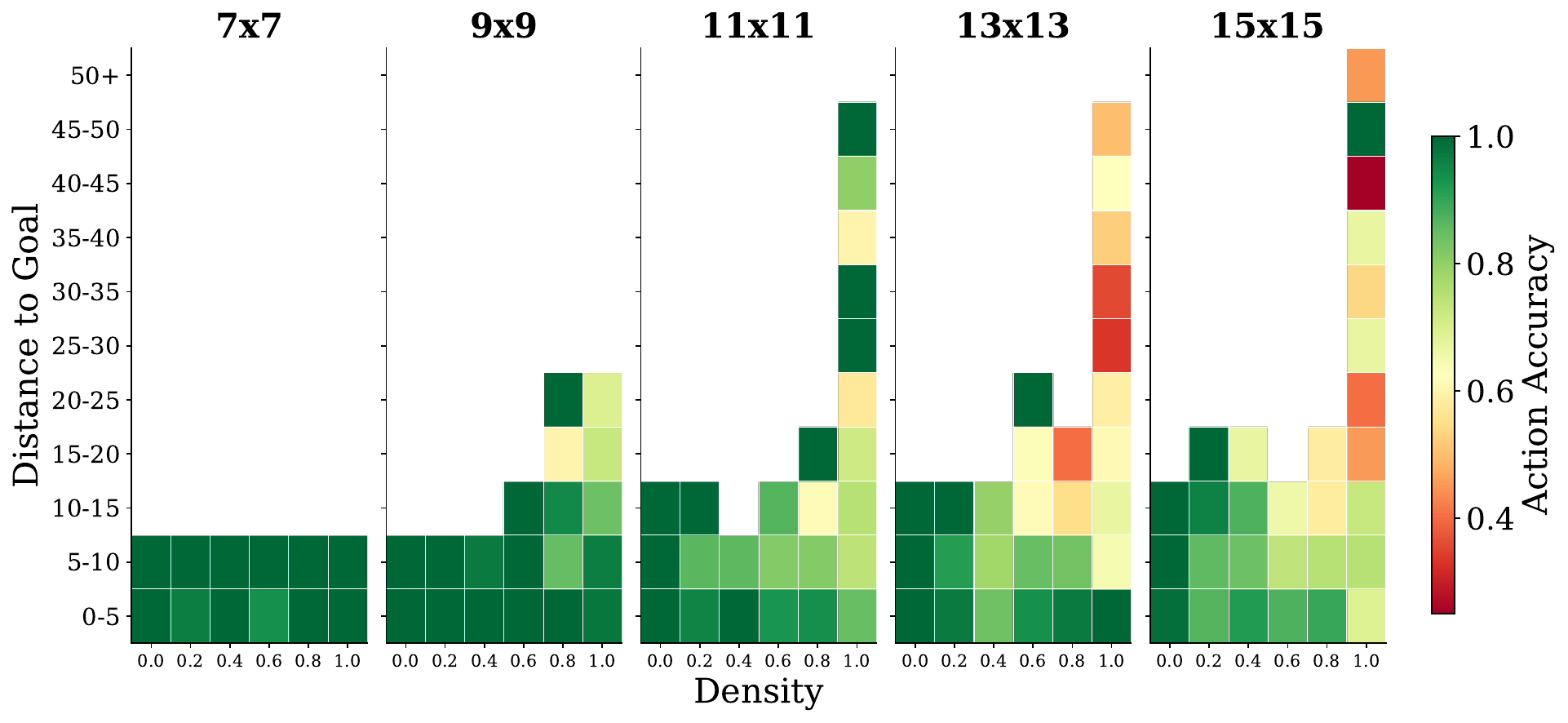}
    \caption{\textbf{Top:} Action accuracy (left) and mean JSD (right) in relation to the agent's distance from the goal. \textbf{Bottom}: Action accuracy by size, complexity, and goal distance.}
    \label{fig:metrics_by_distance}
    \vspace{-0.8em}
\end{figure}

The agent's behavioural metrics vary systematically with task difficulty. Per-action accuracy decreases monotonically with both grid size and obstacle density. 
Conversely, policy entropy and JSD from the optimal policy both increase with grid size and obstacle density.
Together, these results indicate that the agent's policy becomes less optimal and more uncertain on more difficult grids.

We further analyse behavioural metrics as a function of the agent's distance to the goal (\Cref{fig:metrics_by_distance} top). Per-action accuracy decreases linearly with distance for distances below 20 steps, after which the estimates become noisier, and the JSD from the optimal policy increases correspondingly. 
Variance in both metrics grows with distance, indicating less stable behaviour when the agent is farther from the goal. \Cref{fig:metrics_by_distance} (bottom) breaks down per-action accuracy by grid size, obstacle density, and distance to the goal, confirming that increases along any of these three dimensions are associated with lower action accuracy. 
Controlling for the other factors, accuracy decreases most systematically with increasing grid size and obstacle density, with distance to goal only playing a significant role for the larger grid sizes.

\subsection{Robustness to Iso-difficulty Transformations}\label{subsec:iso_dif_experiments}
Having established that the agent's performance varies predictably with task difficulty, we next evaluate its robustness to controlled environment transformations that preserve the difficulty of the navigation task.
We refer to these as \emph{iso-difficulty transformations}.
They allow us to assess whether the agent is biased towards specific grid configurations despite equivalent task structure.
We consider four such transformations, shown in \Cref{fig:example_transforms} (App.~\ref{app:iso_transform_results}): (1) reflection of the grid (\textsc{ReflectEnv}); (2) rotation of the grid by $90^\circ$ (\textsc{RotateEnv}); (3) swapping the agent's start position with the goal position (\textsc{StartGoalSwap}); and (4) transposing the grid (\textsc{TransposeEnv}). Each transformation preserves the grid size, obstacle density, and optimal path length of the original grid, and therefore maintains the difficulty of the task. The optimal policy on the transformed grid is obtained by applying the corresponding transformation to the optimal policy of the baseline grid. 

We apply our transformations to grids $\mathcal{G}$, as defined at the beginning of \Cref{sec:objective_policy_eval}, and compare behavioural metrics between each original grid and its transformed counterpart. For each grid, we compute paired metrics across all trajectories and use a Wilcoxon signed-rank test to assess whether performance differs significantly between baseline and transformed environments. Across all transformations, we find no statistically significant differences in any of the evaluated metrics. 
This indicates that the agent's behaviour in grid worlds is driven by task-relevant information rather than by incidental properties of particular grid configurations. 
Detailed results are reported in App.~\ref{app:iso_transform_results} (\Cref{tab:performance_stats} and \Cref{fig:iso_transforms}).\looseness-1

\begin{figure}[t!]
     \centering
     \begin{subfigure}[b]{0.27\linewidth}
         \centering
         \includegraphics[width=\textwidth]{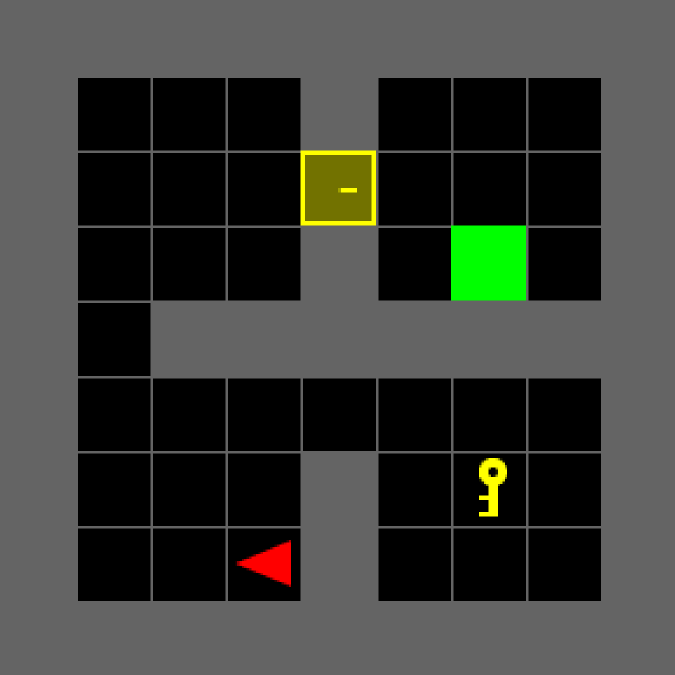}
         \caption*{\textit{KeyDoorEnv}}
         \label{fig:first}
     \end{subfigure}
     \hfill 
     \begin{subfigure}[b]{0.27\linewidth}
         \centering
         \includegraphics[width=\textwidth]{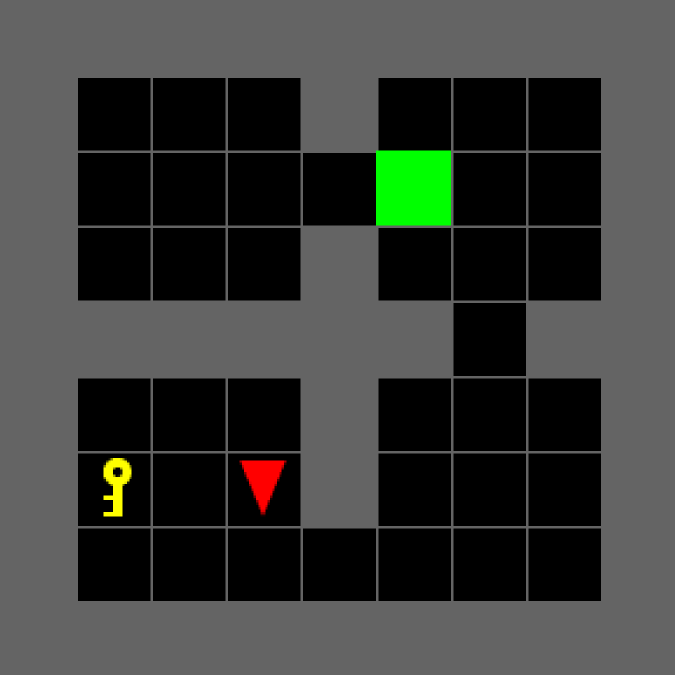}
         \caption*{\textit{KeyNoDoorEnv}}
         \label{fig:second}
     \end{subfigure}
     \hfill
     \begin{subfigure}[b]{0.27\linewidth}
         \centering
         \includegraphics[width=\textwidth]{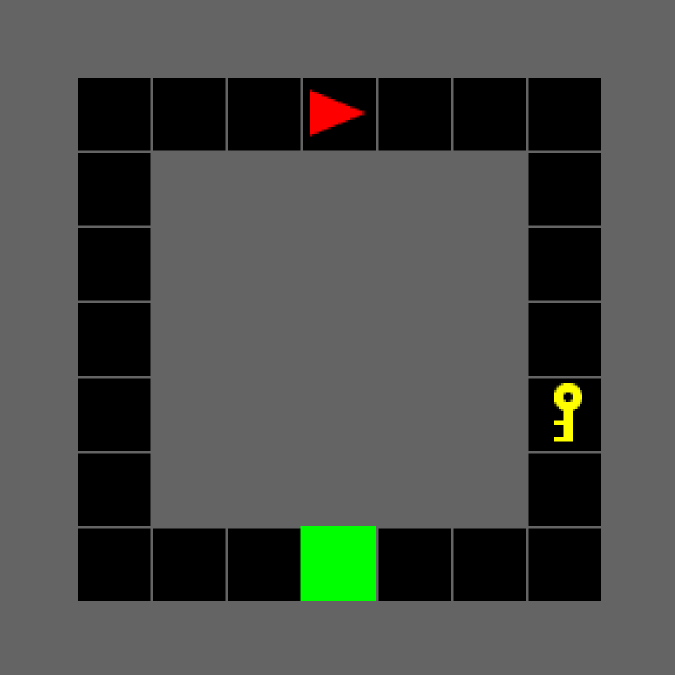}
         \caption*{\textit{2PathKeyEnv}}
         \label{fig:third}
     \end{subfigure}
     \caption{Grid world variants with instrumental and implicit goals. In the text representation, the key and the door are encoded with \texttt{K} and \texttt{D}, and their meaning is explained in the system prompt.}
     \label{fig:key_door_env_variants}
     \vspace{-0.8em}
\end{figure}

\subsection{Instrumental and Implicit Goals}\label{subsec:intrm_impl_goals}
We next examine whether the robustness observed in the previous behavioural evaluations extends to more complex goal structures. We focus on two cases: \textit{instrumental goals}, where an intermediate subtask is necessary for reaching the main task objective, and \textit{implicit goals}, where a goal-like object is present but carries no reward or utility for the task. We instantiate these cases using three grid world variants, shown in \Cref{fig:key_door_env_variants}: \textit{KeyDoorEnv}, \textit{KeyNoDoorEnv}, and \textit{2PathKeyEnv}. The prompt is provided in App.~\ref{sec:prompt_instr_impl}.

\textbf{Instrumental goals.} In \textit{KeyDoorEnv}, the agent must collect a key to unlock a door that blocks the path to the goal. The agent interacts with the key and the door automatically upon reaching the associated cell, with the door acting as a wall if the agent does not have the key.
This setting tests whether the agent can handle a goal structure in which an instrumental subgoal is required to reach the main goal.

\textbf{Implicit goals.} In \textit{KeyNoDoorEnv}, the door is removed, making the key functionally useless. This allows us to test whether a goal-like object influences the agent's behaviour despite its irrelevance to task completion. In \textit{2PathKeyEnv}, the grid contains two optimal paths, one of which contains a vestigial key, allowing us to test whether the agent's path selection is biased towards the key even when collecting it is unnecessary. 
To isolate the effect of the key, we use a matched control grid with identical structure but no key.

We generate $100$ trajectories for \textit{KeyDoorEnv} and \textit{KeyNoDoorEnv}, and $100$ trajectory pairs for \textit{2PathKeyEnv} to account for the key presence. 
We set $T=30$, which is sufficient for solving the maze. Results are summarised in \cref{tab:key_door_results}. 
In \textit{KeyDoorEnv}, the agent achieves a 100\% success rate, correctly collecting the key, unlocking the door, and reaching the goal. Accuracy relative to the optimal policy remains high throughout all subtasks. This indicates successful handling of instrumental goals.
In contrast, performance slightly deteriorates in \textit{KeyNoDoorEnv}, despite the reduced task complexity.
Although the key has no functional utility, the agent still picks it up in 17\% of trajectories. Moreover, among non-optimal actions, 75\% move towards the key, indicating that the key acts as an effective distractor even when it is not ultimately collected.
In \textit{2PathKeyEnv}, the agent is biased towards the key-containing path, with a key pickup rate of 67.3\%. This bias leads to a slight improvement in per-action accuracy (76.0\% vs.\ 74.3\%) but ultimately lowers success compared to the no-key control (71.4\% vs.\ 75.5\%). Trajectories with and without the key also show low Jaccard similarity, indicating substantial behavioural differences induced by the presence of the key.

\begin{table}[t!]
\small
\centering
\caption{Instrumental and implicit goals results.}
\label{tab:key_door_results}
\resizebox{\columnwidth}{!}{
\begin{tabular}{lcc}
\toprule
\multicolumn{3}{c}{\textbf{Results from \textit{KeyDoorEnv} and \textit{KeyNoDoorEnv}}} \\
\midrule
\textbf{Metric} & \textbf{\textit{KeyDoorEnv}} & \textbf{\textit{KeyNoDoorEnv}} \\
\midrule
Success Rate (\%) & $100.0$ & $98.9$ \\
Accuracy (\%) & $98.7 \pm 3.2$ & $97.2 \pm 11.1$ \\
\midrule
\multicolumn{3}{l}{\textit{Stage-specific Accuracy (\%):}} \\
\quad Collecting Key & $98.6 \pm 5.7$ & N/A \\
\quad Opening Door & $99.2 \pm 3.2$ & N/A \\
\quad Reaching Goal & $99.2 \pm 3.3$ & N/A \\
\midrule
\multicolumn{3}{l}{\textit{Key-related Metrics:}} \\
\quad Key Pickup Rate (\%) & $100.0$ & $17.0$ \\
\quad  Key Attraction Bias\textsuperscript{*} (\%) & N/A & $75.0$ \\
\multicolumn{3}{l}{\small \textsuperscript{*}Percentage of non-optimal actions that move towards the key} \\
\midrule
\multicolumn{3}{c}{\textbf{Comparison of Trajectories from \textit{2PathKeyEnv}}} \\
\midrule
\textbf{Metric} & \textbf{With Key} & \textbf{Without Key} \\
\midrule
Success Rate (\%) & $71.4$ & $75.5$ \\
Accuracy (\%) & $76.0 \pm 16.1$ & $74.3 \pm 15.7$ \\
Key Pickup Rate (\%) & $67.3$ & N/A \\
Jaccard Sim. & \multicolumn{2}{c}{$65.6 \pm 35.8$} \\
\bottomrule
\end{tabular}}
\vspace{-1.5em}
\end{table}

In summary, the agent reliably solves tasks with instrumental goals, but is also systematically influenced by goal-like non-functional artefacts. 
We conjecture that this may reflect a conflict between the task goal specified in the prompt and semantic associations acquired during training, such as the common association between keys and progress in games, with the latter not consistently suppressed in favour of the inference-time task goal.

\begin{figure*}[t!]
  \centering
  \begin{minipage}{0.29\textwidth}
    \centering
    \includegraphics[width=\linewidth]{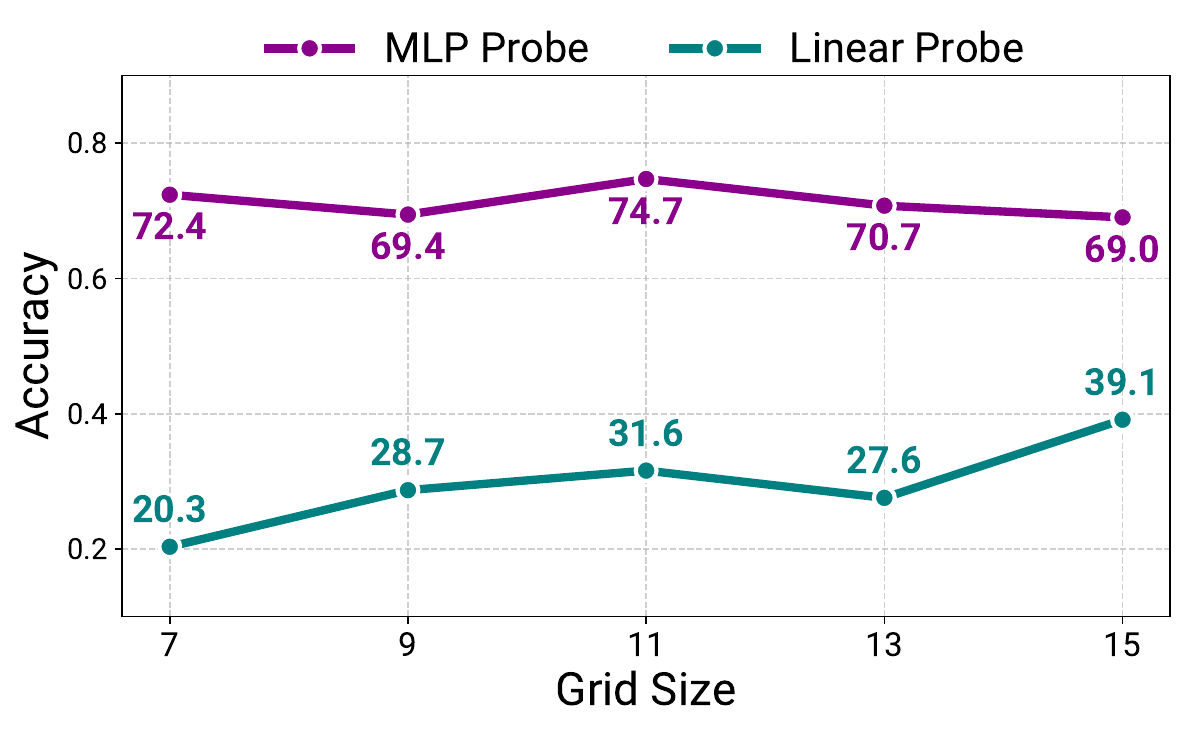}
  \end{minipage}\hfill
  \begin{minipage}{0.39\textwidth}
    \centering
    \includegraphics[width=\linewidth]{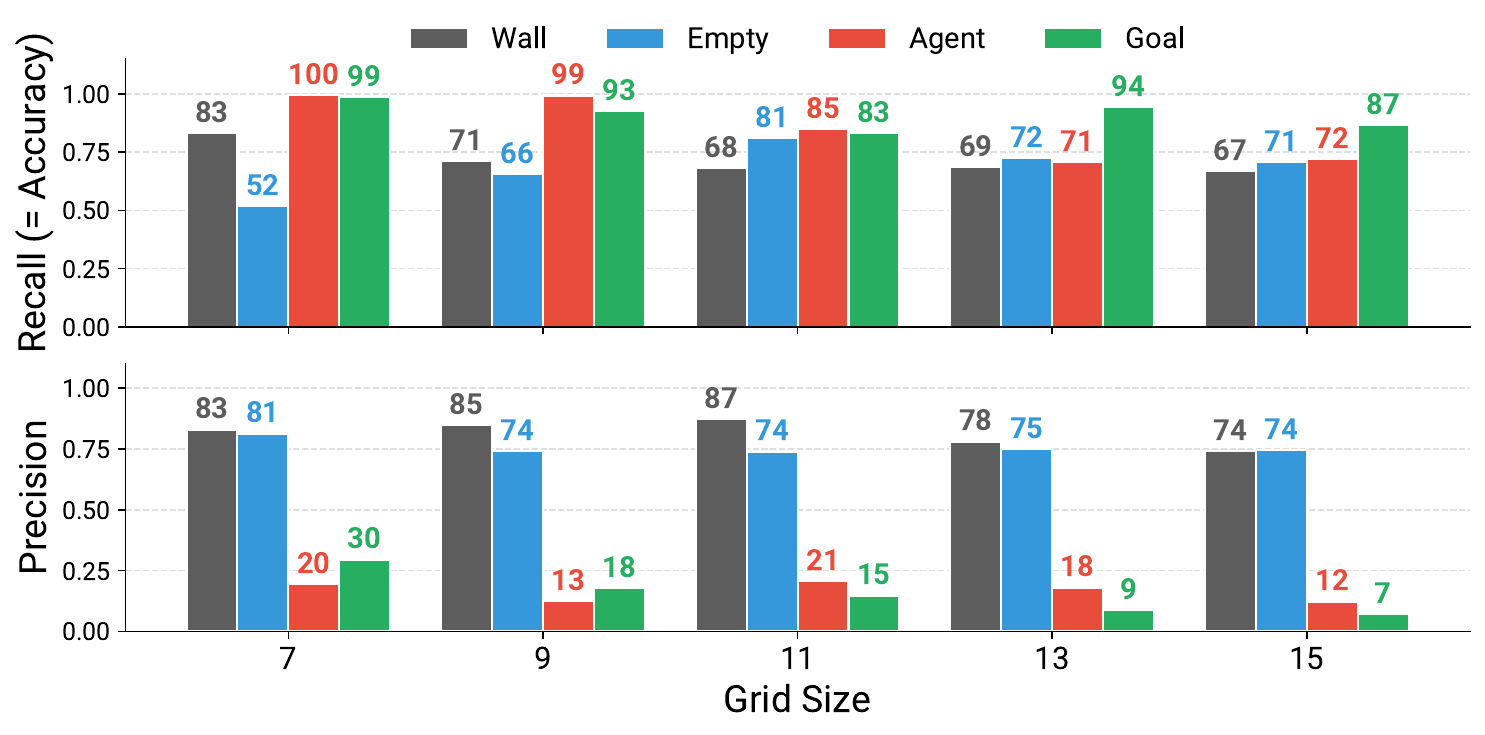}
  \end{minipage}\hfill
  \begin{minipage}{0.29\textwidth}
    \centering
    \includegraphics[width=0.49\linewidth]{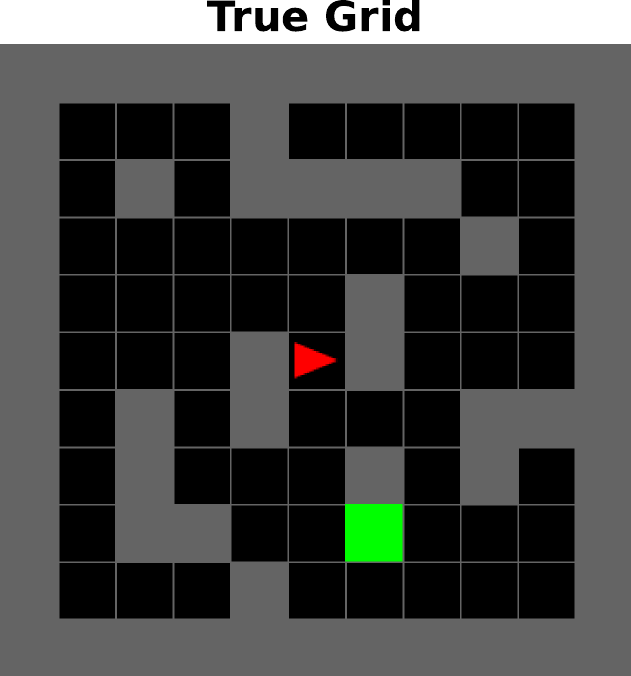}
    \includegraphics[width=0.49\linewidth]{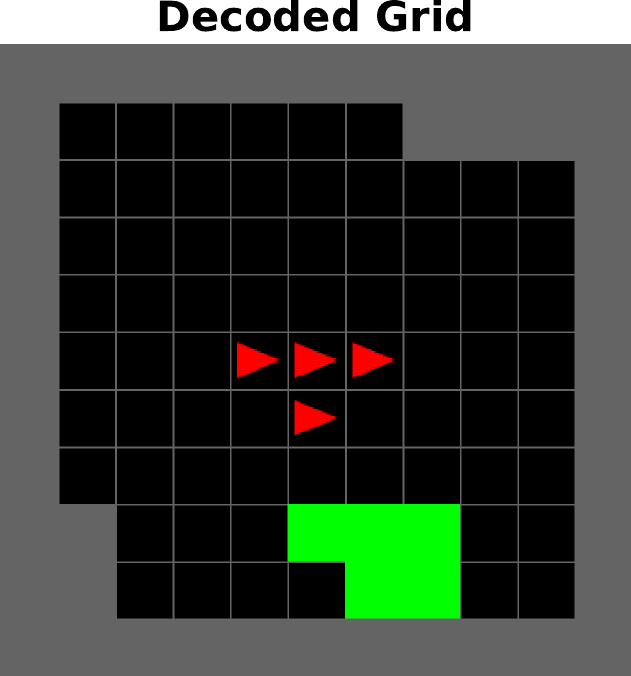}
  \end{minipage}

  \caption{
      Extracting a cognitive map from GPT-OSS-20B representations.
      \underline{Left:} Overall accuracy of an MLP and a linear probe.
      \underline{Centre:} Per-class recall (=accuracy) and precision for varying grid sizes.
      \underline{Right:} A cognitive map decoded from pre-reasoning activations.
  }
  \label{fig:cognitive_maps_overview}
  \vspace{-0.8em}
\end{figure*}

\section{Representational Evaluation}\label{sec:rep-eval}
Behavioural evaluation alone is insufficient to determine an agent's goal-directedness. 
This limitation has been noted in both theoretical~\citep{bellot2025limits} and philosophical work~\citep{rajcicGoalDirectednessEyeBeholder2025,Chalmers_2025}, and has clear practical implications. 
In grid world navigation, for example, an agent may fail to reach the goal state while still acting goal-directedly relative to its own imperfect beliefs about the environment. 
In this section, we therefore analyse the agent's internal world representations to evaluate whether its actions are consistent with those beliefs.

We begin in \Cref{subsec:cognitive_maps} by decoding the agent's beliefs about the environment state, producing what we term \emph{cognitive maps}.\footnote{
    We borrow this term from classic cognitive neuroscience work on navigational tasks \citep{tolman1948cognitive,schmidt2013navigation}.
}  
Building on this, in \Cref{subsec:policies_against_decoded_beliefs}, we evaluate the optimality of the agent's actions with respect to its decoded, subjective cognitive map. 
Finally, in \Cref{subsec:evaluating_plans}, we examine whether goal-directed action plans can be extracted from the agent's internal representations.
In App.~\ref{sec:goal-distance-representation}, we additionally test whether the agent encodes its distance to the goal.

\subsection{Cognitive Maps: Decoding the Agent's Beliefs about its Environment}
\label{subsec:cognitive_maps}

To assess whether the agent's representations encode an internal model of its environment, we prompt GPT-OSS-20B to solve the fully observable grid navigation task described in \Cref{sec:fo_gridworld} and extract residual-stream activations from selected prompt positions. Specifically, we use the final three pre- and post-reasoning tokens in the model chat template (\texttt{<|end|>}, \texttt{<|start|>}, and \texttt{assistant}), extracting activations at layers 7, 15, and 23.
Loosely inspired by \citet{li2023emergent}, we construct training examples by augmenting each activation with the $(x, y)$ coordinates of the queried cell. This yields input--output pairs of the form $([\text{act}, x, y], c)$, where $c \in \{\agent, \goal, \wall, \open, \padding\}$.\footnote{When training general probes, grids smaller than $15 \times 15$ are padded by assigning positions outside the original grid the \padding{} label. After constructing the full dataset across all grid sizes, we upsample minority classes to match the size of the largest class.\looseness-1} We then train linear and MLP classifiers on the resulting pairs to decode cell types across grid positions. For the MLP probe, we use a two-layer architecture with ReLU activation and a hidden dimension of $1024$. Both probes are trained using an AdamW optimiser with weight decay, and normalisation is applied before training. 
At test time, we apply the probes to each grid coordinate and reconstruct the model's \textit{cognitive map} by selecting $\arg \max_c P(\,c\,| \,\text{act}, x, y \,)$ at each position.
This cognitive map represents the model's decoded belief over the current grid state.
We train \textit{general} and \textit{size-specific} variants of these probes using grids of sizes $7,9,11,13,15$. In the general case, we pad smaller grids to size $15$ by assigning $c=$ \padding{} to missing positions. Unless otherwise stated, the following experiments focus on general probes trained on pre-reasoning activations from the intermediate layer 15. Additional results across layers, grid sizes, and goal-distance probes are reported in App.~\ref{sec:cogmap-across-sizes}.

\begin{figure}[t!]
  \centering
  \includegraphics[width=\linewidth]{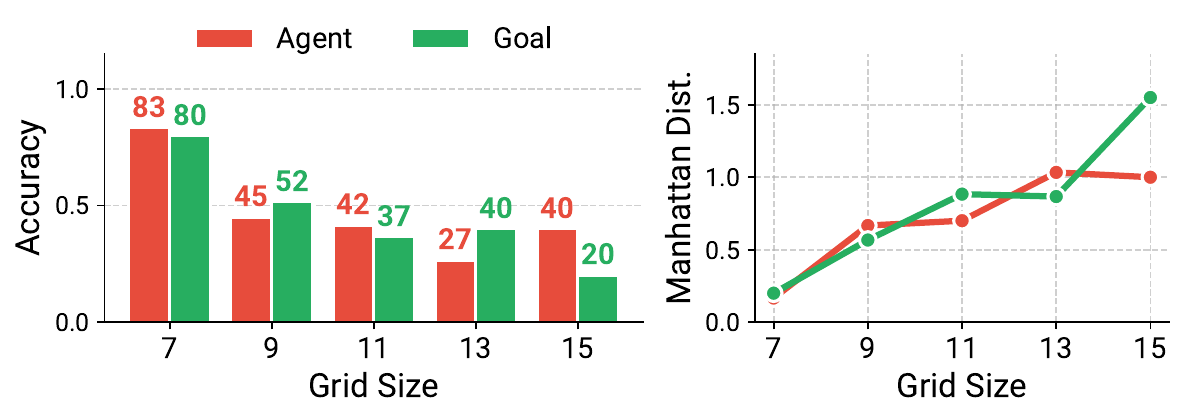}
  \caption{Probe performance for locating agent and goal positions. Binary localisation accuracy drops as the grid size increases, but avg. Manhattan distance to true locations remains bounded.}
  \label{fig:highest_prob_metrics}
\vspace{-0.8em}  
\end{figure}

\textbf{How is the Environment Encoded in Model Activations?}
\Cref{fig:cognitive_maps_overview} (left) shows the accuracy of cognitive maps reconstructed by linear and MLP probes across various grid sizes. The MLP probe decodes cell identities with around 70\% accuracy, reaching a maximum of 75.7\% for $11\times11$ grids. Linear probes underperform at 39.1\% accuracy in the same setting, suggesting that environment information is encoded non-linearly in model representations. \Cref{fig:cognitive_maps_overview} (centre) presents a per-class performance breakdown of the MLP probe in terms of its recall (top) and precision (bottom). Recall scores show that all cell identities are retrieved robustly across grid sizes, with especially high recall for goal (83-99\%) and agent (72-100\%) positions. We note that probes assign \agent{} and \goal{} labels to multiple cells in the neighbourhood of their respective true locations, resulting in the high recall but low precision trend discussed earlier (\Cref{fig:cognitive_maps_overview}, right).
In contrast, information about the position of walls is not represented in detail, as reflected by the lower recall for the \wall{} class. These results suggest that the agent's internal representations encode a coarse spatial map of the environment, preserving approximate task-relevant information about agent position and goal location.

\textbf{Locating Agent and Goal.}
Given the coarse localisation of \agent{} and \goal{} cells in cognitive maps, we set out to test how accurately the true agent and goal locations can be recovered. 
We measure top-1 accuracy using the grid coordinates with the highest predicted $P(c = \text{\agent})$ and $P(c = \text{\goal})$, with results shown in \Cref{fig:highest_prob_metrics}.
We find that agent and goal localisation accuracy decreases steadily with grid size. 
However, the Manhattan distance between predicted and true locations remains lower than 2 even for large $15\times15$ grids, suggesting that information about agent and goal positions is encoded coarsely in proximity to their true locations.\looseness-1

\begin{figure}[t!]
  \centering
  \includegraphics[width=\linewidth]{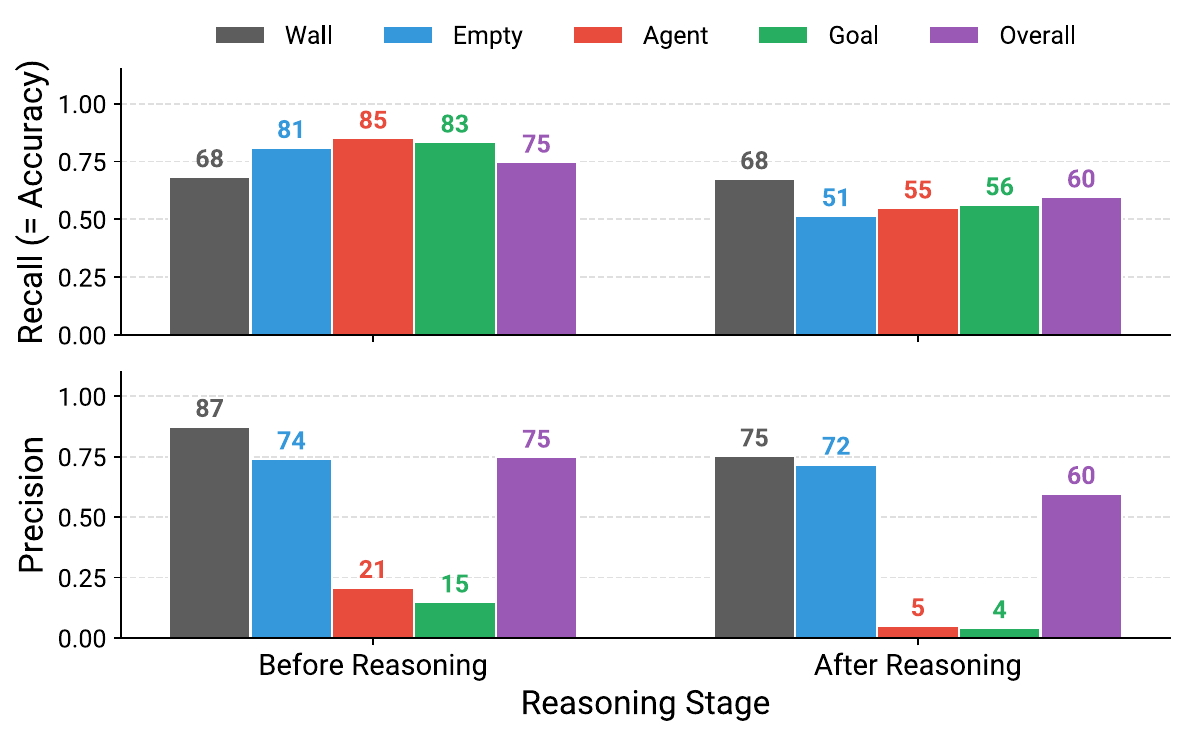}
  \caption{Performance of cognitive map before and after reasoning. Cognitive map accuracy drops significantly after reasoning.} 
  \label{fig:post_reasoning_metrics}
\vspace{-0.5em}
\end{figure}

\textbf{Goal Location Information Degrades after Reasoning.}
We next evaluate how decoded cognitive maps change from pre- to post-reasoning activations.
As shown in \Cref{fig:post_reasoning_metrics}, overall probe accuracy drops from 75\% before reasoning to 60\% after reasoning, with a notable decrease  in \agent{}, \goal{} and \open{} recall, as well as in \agent{} and \goal{} precision.
These results indicate that the coarse environment state information recovered by the cognitive map probes is substantially less available after reasoning.
This may reflect a reorganisation of the model's representations, in which spatial information becomes less directly decodable as the model shifts towards encoding information more directly relevant to action selection.\looseness-1

\subsection{Evaluating Policies against Decoded Beliefs}
\label{subsec:policies_against_decoded_beliefs}
Behavioural deviations from the optimal policy do not, by themselves, show that the agent is not acting goal-directedly. If the agent's internal representation of the environment is incomplete or inaccurate, an action may be suboptimal in the true grid while still being appropriate relative to the agent's own beliefs. 
We therefore test whether the agent's behaviour is consistent with its decoded cognitive maps, focusing in particular on cases where its actions deviate from the ground-truth optimal policy.
To do so, we compare the agent's observed action sequence with the optimal policy derived from the cognitive map decoded at each trajectory step using the general MLP probe with layer-15 pre-reasoning activations. 
As in \Cref{subsec:cognitive_maps} (and \Cref{fig:highest_prob_metrics}), we first identify a single \agent{} and \goal{} location by selecting the grid cells with the highest predicted probabilities for the corresponding classes. 
We then compute optimal actions on both the decoded and ground truth grids.
\Cref{tab:results_behavior_in_decoded_grid} reports the accuracy of the agent's actions with respect to optimal policies defined on the decoded cognitive map (\textit{Acc.\ Dec.}) and the ground truth grid (\textit{Acc.\ GT}), as well as the fraction of actions that are optimal under both policies (\textit{Agreem.}).

\textit{Acc.\ Dec.}, the accuracy relative to the optimal policy in the decoded cognitive map, decreases with grid density but is high across grid sizes (average: 82.5\%).
This indicates that the agent's actions are broadly consistent with its internal world representation.
Of particular interest is the recovery metric \textit{Rec.}, which is the proportion of actions that are suboptimal in the true environment but optimal given the agent's decoded cognitive map (see App.~\ref{sec:app-inverse-recovery-rate} for the inverse recovery rate).
\textit{Rec.}~ranges between 37.4\% and 88.4\% (average: 57.9\%), indicating that a substantial fraction of failures can be attributed to inaccurate or fuzzy world representations rather than a lack of goal-directedness, particularly in low-density and medium-to-large grids.

While the agreement between the policy defined on the decoded vs.\ ground truth grid is high across conditions (\textit{Agreem.}\ averages 83.9\% and is always above 77\% except for the highest density grids),
we remark that \textit{Acc.\ Dec.}\ is consistently lower than \textit{Acc.\ GT}. 
This may be due to the fact that we derive a single location for the agent and goal by selecting the grid cell with the highest predicted probability. 
This approach does not capture the uncertainty evident in the decoded maps, which exhibit blurred agent and goal spatial representations, as discussed in \Cref{subsec:cognitive_maps} and shown in \Cref{fig:cognitive_maps_overview}. 
As a result, actions may be optimal with respect to nearby cells rather than the single $\mathrm{argmax}$ location.

\textbf{Uncertainty-Aware Cognitive Maps.} 
\begin{figure}[t!]
  \centering
  \includegraphics[width=\linewidth]{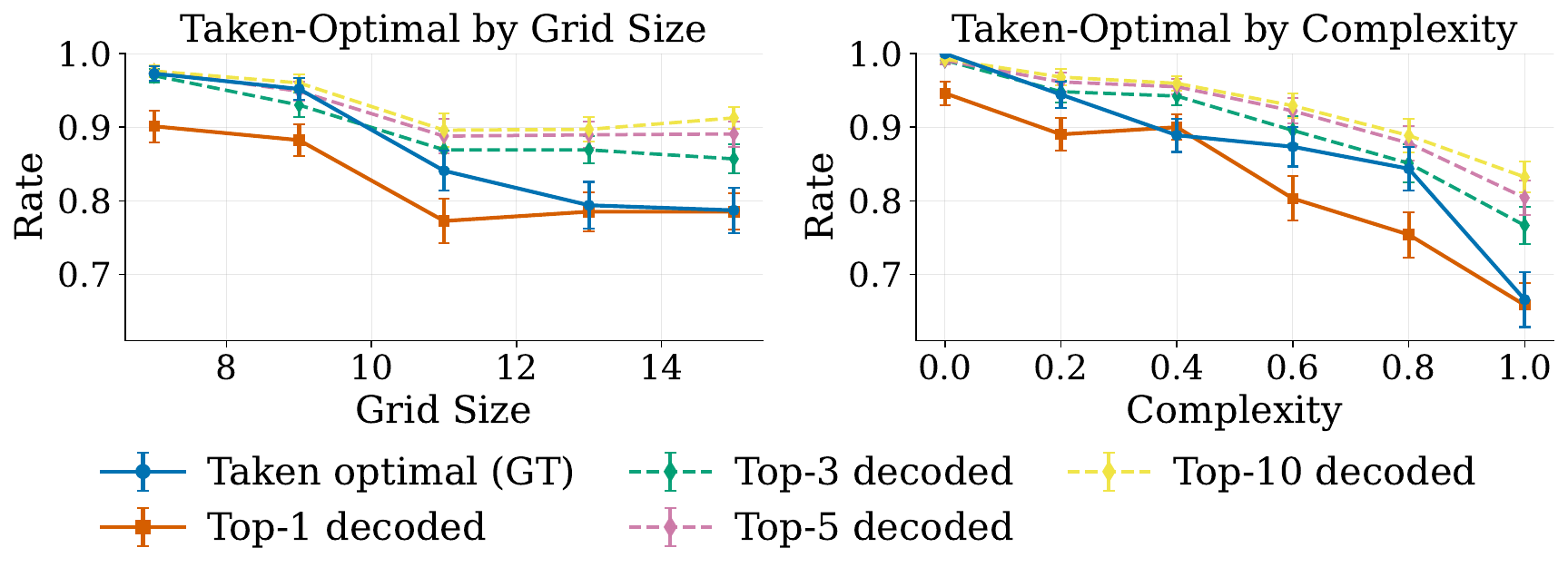}
  \caption{Per-action accuracy against the optimal policy defined with respect to the top-$k$ decoded grids.} 
  \label{fig:decoded_optimality}
\vspace{-0.5em}
\end{figure}
To account for the agent's fuzzy world representations and mitigate the low precision of goal and agent location probes, we consider top-$k$ decodings of the cognitive map, evaluating action optimality with respect to a set of $k$ candidate agent and goal positions per grid.
\cref{fig:decoded_optimality} shows the proportion of optimal actions for the ground truth grid, top-1 decoded grid (results from \cref{tab:results_behavior_in_decoded_grid}), and top-$k$ decoded grids for $k\in\{3, 5, 10\}$. 
Accuracy against top-$k$ decodings is consistently higher than accuracy against the ground truth grid across grid sizes and complexity bins.
This indicates that accounting for uncertainty in the probe predictions allows the decoded grids to explain a larger share of the agent's actions.
We interpret these results as evidence that top-1 decoding collapses the agent's representational uncertainty. 
As the state space grows and the agent's internal representations become fuzzier, its actions are better characterised as planning under a distribution over plausible states.

\textbf{Intervention via Activation Patching.} Finally, to complement our observational probing analyses, we conducted preliminary activation patching experiments on the residual stream of GPT-OSS-20B, using corrupted examples that move either the \agent{} or the \goal{} square. We find that patching changes the model's action distribution only when applied across all layers simultaneously, and only at grid state tokens or the token immediately preceding the action output. Single-layer interventions applied to the chat template tokens proved ineffective (see App.~\ref{app:activation-patching}). 
This asymmetry between readouts and interventions is consistent with recent findings suggesting that prompt-related attributes can be read by probing classifiers at various specific points in the forward pass---often from individual layers---but  interventions require more distributed representation editing to achieve a measurable impact \citep{choi2025scalably}. Identifying layer- and position-specific intervention strategies for our setup is thus an important future direction.

\subsection{Evaluating Plans}
\label{subsec:evaluating_plans}

In this section, we examine whether the agent's internal representations encode goal-directed planning information, and how this encoding differs before vs.\ after reasoning.
We consider the same single-goal navigation task as in previous sections.
For each trajectory, we extract residual-stream activations from layers $\ell \in \{7, 15, 23\}$ at two stages: (i)~\textit{pre-reasoning}, using the final prompt tokens immediately before the model begins reasoning, and (ii)~\textit{post-reasoning}, using the final reasoning tokens immediately before the model outputs its first action.
Each example is labelled with a target action sequence $\mathbf{a}_{1:T} = (a_t)_{t=1}^{T}$, derived from the executed actions in the trajectory $\tau^{\policy}(s_0) = (s_t, a_t)_{t=0}^{T}$ for a given grid instance, with $T = 10$.
We train the plan decoder on 3{,}000 trajectories, use a 600-trajectory validation set, and report results on the 300-trajectory test set used in~\Cref{sec:objective_policy_eval}.
Trajectories are sampled from grid sizes 7--15 with varying complexities and start/goal configurations, yielding a diverse distribution of path lengths and planning difficulty.

\begin{table}[t!]
\vspace{0.3em}
\small
\centering
\caption{Policy evaluation against decoded beliefs. \textit{Acc.\ GT}: action accuracy w.r.t.\ the optimal policy on the ground truth grid; \textit{Acc.\ Dec.}: action accuracy w.r.t.\ the optimal policy on the cognitive map; \textit{Agreement}: \% optimal actions for both policies; \textit{Recovery}: \% optimal actions only for the cognitive map. 
Grid size results (top) are averaged across density, and vice versa (bottom).}
\label{tab:results_behavior_in_decoded_grid}
\resizebox{\columnwidth}{!}{
\begin{tabular}{ccccc}
\toprule
$\boldsymbol{n}$ & \textbf{Acc. GT (\%)} & \textbf{Acc. Dec. (\%)} & \textbf{Agreement (\%)} & \textbf{Recovery (\%)} \\
\midrule
7 & $97.3$ & $90.1$ & $91.5$ & $48.3$ \\
9 & $95.2$ & $88.2$ & $88.3$ & $42.4$ \\
11 & $84.1$ & $77.3$ & $81.7$ & $74.2$ \\
13 & $79.4$ & $78.5$ & $79.7$ & $64.5$ \\
15 & $78.7$ & $78.5$ & $78.4$ & $60.1$ \\
\midrule
$\boldsymbol{d}$ & \textbf{Acc. GT (\%)} & \textbf{Acc. Dec. (\%)} & \textbf{Agreement (\%)} & \textbf{Recovery (\%)} \\
\midrule
$0.0$ & $100.0$ & $94.6$ & $98.6$ & N/A \\
$0.2$ & $94.5$ & $89.1$ & $93.2$ & $88.4$ \\
$0.4$ & $88.9$ & $90.0$ & $93.7$ & $82.4$ \\
$0.6$ & $87.3$ & $80.3$ & $83.2$ & $49.6$ \\
$0.8$ & $84.4$ & $75.4$ & $77.3$ & $47.8$ \\
$1.0$ & $66.6$ & $65.8$ & $57.6$ & $37.4$ \\
\bottomrule
\end{tabular}}
\vspace{-0.8em}
\end{table}

\textbf{Plan Decoder Architecture.}
Our goal is to decode the entire plan from a fixed set of activations, while minimising any additional planning or inference performed by the probe.
Let $\mathbf{h}_1,\mathbf{h}_2,\mathbf{h}_3 \in \mathbb{R}^{2880}$ denote the three extracted token activations at a chosen layer and stage.
We first map each~$\mathbf{h}_i$ through a shared bottleneck consisting of a linear projection to 1024 dimensions followed by LayerNorm:
$\tilde{\mathbf{h}}_i = \mathrm{LN}(W \mathbf{h}_i) \in \mathbb{R}^{1024}$.
We then decode a horizon-$T$ plan using a Transformer decoder with $T$ learned query embeddings $\mathbf{q}_1,\ldots,\mathbf{q}_{T}$.
Each query corresponds to a plan step index and performs cross-attention over the same set of token activations $\{\tilde{\mathbf{h}}_1,\tilde{\mathbf{h}}_2,\tilde{\mathbf{h}}_3\}$.
We evaluate Transformer variants with 1, 2, and 4 layers, each with 8 attention heads per layer.
A final linear head produces a distribution over actions for each step $t \in [T]$,
$p(a_t \mid \tilde{\mathbf{h}}_{1:3}) = \mathrm{softmax}(W_o \mathbf{z}_t)$, where $\mathbf{z}_t$ is the decoder output at query slot $t$. 

\begin{figure}[t!]
    \centering
    \includegraphics[width=0.95\linewidth]{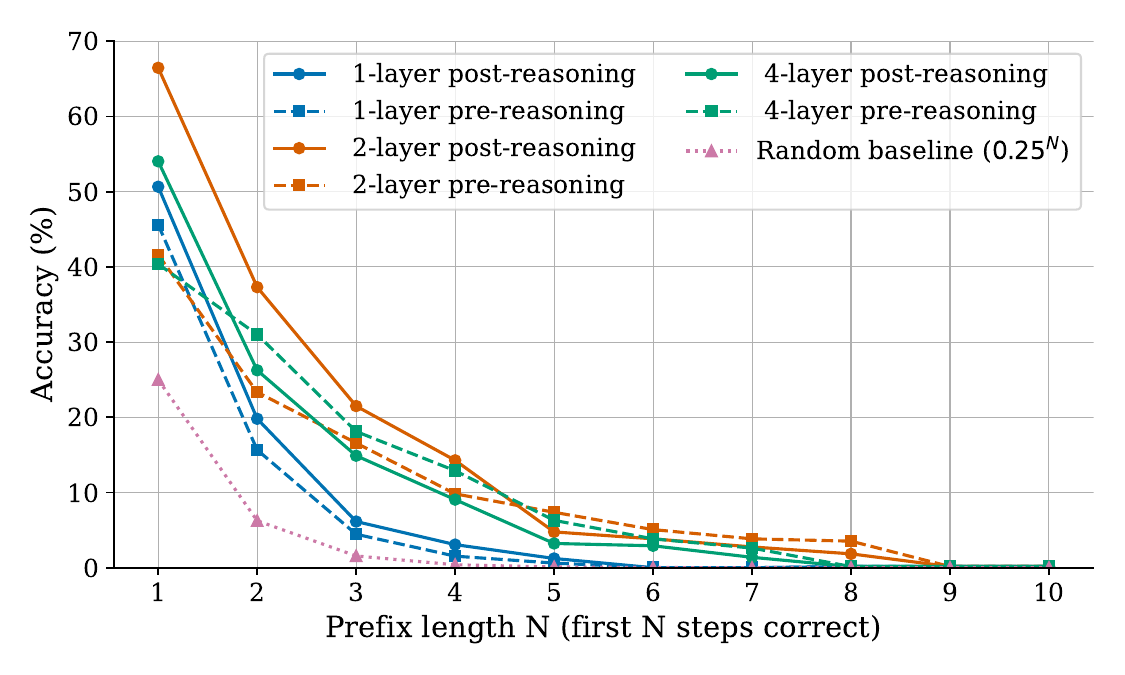}
    \caption{Prefix accuracy of one-shot plan decoding across decoder capacities. We compare Transformer decoders with 1, 2, and 4 layers, with activations extracted before vs.\ after reasoning.}
    \label{fig:prefix_accuracy}
    \vspace{-0.8em}
\end{figure}

Importantly, we design the decoder to predict the entire plan $\hat{\mathbf{a}}_{1:T}$ simultaneously from the input activations, rather than predicting $\hat{a}_t$ autoregressively conditioned on previously decoded actions $\hat{a}_{<t}$. Autoregressive decoding would introduce an additional channel for the probe to create plan structure: early predicted actions can implicitly constrain later actions via simple continuation heuristics, even if the underlying representations only weakly specify a full-horizon trajectory. 
By contrast, in one-shot decoding, later steps cannot condition on earlier predictions, so coherent multi-step structure in 
$\hat{\mathbf{a}}_{1:T}$  must be supported by information already present in the base model's activations. 
Thus, accuracy above baseline under one-shot decoding is more diagnostic of plan information encoded in the model's representations than of computation performed by the probe itself.
We evaluate plan decodability using \emph{prefix accuracy}, defined as the fraction of episodes for which the first $N$ predicted actions exactly match the target plan prefix.
For a predicted plan $\hat{\mathbf{a}}_{1:T}$ and target action sequence $\mathbf{a}_{1:T}$ (with $T=10$), prefix accuracy at $N$ is $\Pr[\hat{\mathbf{a}}_{1:N}=\mathbf{a}_{1:N}]$.
We report prefix accuracy for $N \in \{1,\ldots,10\}$ for both pre-reasoning and post-reasoning activations.
As a baseline, random guessing among four actions yields $0.25^N$.

\textbf{Results.} \Cref{fig:prefix_accuracy} shows that all probes exceed the $0.25^N$ baseline at short horizons, indicating that GPT-OSS-20B activations encode non-trivial information about upcoming action sequences.
Across decoder capacities, the 2-layer probe performs best overall, particularly with post-reasoning activations: it achieves 66.49\% accuracy at $N=1$ and remains strongest through $N=4$.
For longer prefixes, the 2-layer pre-reasoning probe has the strongest performance, reaching 7.3\% at $N=5$, 5.0\% at $N=6$, and 3.8\% at $N=7$.
The 1-layer probe also retains substantial one-step decodability (45.5\% pre-reasoning, 50.6\% post-reasoning), but deteriorates rapidly with prefix length, while the 4-layer probe achieves non-trivial longer-prefix accuracy without consistently improving over the smaller probes.
Varying probe capacity provides a control against the possibility that the decoder itself is solving the navigation task.
If performance were driven by decoder-side navigation, larger decoders should consistently outperform smaller ones.
Instead, performance is non-monotonic in decoder capacity: the 2-layer probe outperforms the 4-layer probe post-reasoning, and even the 1-layer probe recovers strong one-step planning information.
This suggests that immediate action information is readily accessible to a low-capacity readout, while multi-step plan structure requires additional capacity to extract.

Focusing on the 2-layer probe as the strongest overall readout, we observed that post-reasoning activations improve one-step decoding relative to pre-reasoning (66.49\% vs.\ 41.5\% at $N=1$), consistent with reasoning increasing the separability of the next action (see~\Cref{subsec:cognitive_maps}).
This advantage persists through $N=4$, while for longer prefixes pre-reasoning activations become more decodable, crossing at $N=5$.
This crossing point suggests again a trade-off induced by reasoning, with post-reasoning activations more strongly supporting \emph{local} action selection, and pre-reasoning representations preserving more recoverable long-horizon trajectory structure. 
One interpretation of this trade-off is that reasoning shifts representational emphasis from broader environment-related structure towards near-term action selection.
Finally, additional analyses (reported in App.~\ref{sec:app-plan-decoder-probe}) show that the decoder does not recover the exact trajectory length in most cases, but it reliably predicts a close estimate.
Since trajectory length in this setting is tightly coupled to progress towards the goal under the executed policy, this provides complementary evidence that the model's internal states encode coarse plans beyond the next action.

\section{Conclusion}\label{sec:conclusion}

We have presented a framework for analysing the goal-directedness of LLM-based agents that integrates behavioural evaluation with representation probing and demonstrated its utility in a grid world navigation setting with GPT-OSS-20B as the agent under study.
Behaviourally, the agent shows systematic sensitivity to task difficulty while remaining robust to goal-irrelevant environmental variations, providing initial evidence of goal-directedness. The agent also succeeds in multi-goal grid worlds with instrumental subgoals, though its behaviour is biased by goal-like but task-irrelevant cues.
Representational analyses uncovered structured internal states consistent with an interpretation of the agent as goal-directed. 
The model encodes cognitive maps that capture task-relevant spatial information about the location of the agent and the goal, and its representations exhibit a shift across the reasoning process: pre-reasoning, they preserve spatial information about the environment and longer-horizon plans, while post-reasoning, they have a narrower focus on next action selection.
This finding suggests that reasoning reorganises information to support effective control.\looseness-1

Our controlled setup enables precise measurement but abstracts away from the complexity of real-world agentic settings, making extension to more complex environments an important direction for future work. 
Moreover, while we find consistent relationships between internal representations and behaviour, simple activation patching does not reliably alter behaviour, leaving the establishment of causal links as an open challenge.
In this context, probing alternative grid encodings \citep[see, e.g.,][]{ivanitskiy2023structured} may reveal aspects of the environment state not captured by our probes.
Addressing these questions, and extending the framework across architectures, scales, and training regimes, will be important for assessing the generality of our findings.

Looking forward, the methods and insights from this work provide a foundation for developing more comprehensive approaches to goal attribution and monitoring in agentic systems. The development of rigorous approaches to evaluating goal-directedness is a prerequisite for making high-confidence claims about agents' goals and potential related risks, and for informing the responsible deployment and oversight of increasingly autonomous AI systems.

\section*{Acknowledgements}
This project was supported by \href{https://sparai.org}{SPAR}. GS acknowledges support by the NDIF project (U.S. NSF Award IIS-2408455). We thank Cozmin Ududec, Dima Krasheninnikov, Gonçalo Guiomar, Michael Hanna, and members of the BauLab at Northeastern University for helpful discussions. 

\section*{Author Contributions}
\label{sec:author-statement}
RA, FC, CM, GS, and MG conducted the literature review of \Cref{sec:related_work}, identified and synthesised relevant prior work across research paradigms, and developed the framing that situates the project's contributions within existing research on goal-directedness.
RA, ND, and AN worked on implementation of code infrastructure for \Cref{sec:objective_policy_eval}. ND developed the iso-difficulty transformations and conducted the analysis for \Cref{subsec:baseline_tasks}, \Cref{subsec:iso_dif_experiments}, App.~\ref{app:po_appendix}, App.~\ref{app:additional_objective_results}, and App.~\ref{app:iso_transform_results}. AN developed the procedures for generating the base environments and the variants with the key, and conducted the analysis for \Cref{subsec:intrm_impl_goals}. GS advised the design of representational evaluation experiments of \Cref{sec:rep-eval}, developed a unified pipeline for trace generation, activation extraction, and probe training, and built a trace viewer to explore probing results. EK designed, implemented, and evaluated the cognitive map probes in \Cref{subsec:cognitive_maps} and App.~\ref{app:additional-representational-results}. RA, ND, and AN conducted the analysis for \Cref{subsec:policies_against_decoded_beliefs}. MN designed, implemented, and evaluated the plan decoder in \Cref{subsec:evaluating_plans}.  MG conceived and led the project, provided ongoing scientific guidance, and coordinated the project's execution. 
All authors contributed to the conceptualisation of the study and the preparation of the manuscript.

\section*{Impact Statement}
Goal-directedness lies at the core of agency: understanding whether an artificial system pursues goals, which goals it pursues, and how those goals relate to its behaviour is foundational for explaining, predicting, and governing advanced AI systems. Progress on this problem has broad scientific significance, spanning cognitive science, linguistics, neuroscience, philosophy of action, macroeconomics, and decision theory, but it also carries important practical implications as AI systems are increasingly deployed in autonomous, long-horizon, and high-stakes settings.

The ability to reliably evaluate goal-directedness is particularly critical from a safety perspective.
As AI systems become more capable, behavioural success alone becomes an increasingly weak signal of alignment. 
Systems driven by misaligned internal objectives may nonetheless appear competent, compliant, or even helpful because aligned behaviour is instrumentally useful. 
Recent work has highlighted risks such as alignment faking and sandbagging \cite{greenblatt2024faking, vanderweij2025sandbagging, taylor2025auditing}, with models behaving deceptively or underperforming during evaluation, arguably to avoid modification or oversight. 
While claims of deliberate ``AI scheming'' would be highly consequential if substantiated, current evidence remains limited, often anecdotal, and difficult to interpret without clear hypotheses, controls, or mechanistic grounding \cite{summerfield2025lessons}. This paper takes a first step towards addressing these limitations.

\bibliography{telos}
\bibliographystyle{icml2026}


\clearpage
\appendix

\section{Partially Observable Grid World}\label{app:po_appendix}
We also examined a partially observable Grid World, wherein the agent can see the full grid but many of the cells are hidden with fog. Once the agent ``sees'' a cell, it is permanently revealed. An agent sees cells around itself with a visibility radius of 3 cells. An example of what the agent sees is given in \cref{fig:po_grid}. The seen radius is blocked by walls; specifically, we use Bresenham's line algorithm to trace lines on the grid.

\begin{figure}[!ht]
    \centering
    \begin{BVerbatim}
  0 1 2 3 4 5 6 
0 * * # # # # # 
1 * * _ _ # _ # 
2 * * # _ # _ # 
3 * * # _ _ A # 
4 * * # # # # # 
5 * * * * * * * 
6 * * * * * * * 
    \end{BVerbatim}
    \caption{Partially Observable representation of a grid. The agent is at "A" and the goal (unseen) will be represented by "G". Hidden spaces are represented by "*", while revealed spaces are represented by "\_". }
    \label{fig:po_grid}
\end{figure}

This setting is formalised as a Partially Observable Markov Decision Process. A POMDP is a 7-tuple $(\states, \actions, \transitions, \reward, \obs, \obsfunc, \gamma)$, which includes all elements of an MDP plus:
\begin{itemize}
    \item $\obs$: A finite set of observations the agent can receive.
    \item $\obsfunc(o | s', a) = \mathbb{P}(o_t=o|s_t=s', a_{t-1}=a)$: The observation function, which gives the probability of receiving observation $o$ after taking action $a$ and landing in the (hidden) state $s'$.
\end{itemize}

We provided the model with a history of past state--action tuples as well as the full conversation history.
We initially planned on running the same set of ablations as the fully observable case: grid sizes, complexity levels, and iso-difficulty transforms. 

\subsection{Difficulties with Partial Observability}

Upon initial testing with partial observability, we discovered that  GPT-OSS-20B has significant difficulties in the partially observable case. We observed several undesirable behaviours, including redundant backtracking, running into walls, and moving towards known dead-ends. 
Increasing the reasoning effort to ``high'' did not alleviate these issues. 
Although it is non-standard, we even tried providing the reasoning step of the model alongside the standard output for each step in the conversation history---this did not help. 
Qualitative inspection of the reasoning chain revealed that the model usually focuses almost exclusively on what its next step should be based on where it currently is, and almost never reasons about its past actions. Even for a more powerful model, GPT-OSS-120B, we rarely observed it talk about ``backtracking'' when on ``high'' reasoning.

In order to determine whether or not this was a capability issue of the models, we performed the same tests with a frontier model, GPT-5.1-Thinking. 
Even this frontier model displayed the same suboptimal behaviours of redundant backtracking and moving towards known dead-ends (although we did not observe it try to run into walls). We also observed the same pathological reasoning wherein the model focused on the current state (``tunnel vision'') and  failed to reason about past actions. 

We hypothesise that these consistent failures in the partial observability case are due to models not being trained on similar problems, and to training on reasoning problems (e.g., math or coding) not generalising to maze navigation---similar to other out-of-distribution reasoning problems, like the ``Alice in Wonderland'' problem discussed by \citet{nezhurina2025alicewonderlandsimpletasks}.
Due to these persistent issues, we decided to leave partial observability to future work. 

\section{Behavioural Evaluation Metrics}
\label{app:behavioural-eval-metrics}

\subsection{Capability Metrics} We evaluate agent performance with various metrics, defined below. The \emph{goal success rate} (GSR) is defined as the expected fraction of trajectories that terminate at the goal state:
\begin{align}
\mathrm{GSR}
\;\coloneq\;
\mathbb{E}_{\tau\sim\pi}
\left[
GS(\tau(s_0))
\right],
\end{align}
where $
GS(\tau(s_0))
\;\coloneq\;
\mathbbm{1}\!\left(s_T=\goalstate\right)
$. In practice, we sample a finite number of trajectories used to compute the GSR, as well as for the other metrics below.

We evaluate the agent's adherence to the optimal policy using a two-step accuracy metric. First, we define the per-action accuracy for a single trajectory $\tau_i$ of length $T_i$ as:

\begin{align}
    \text{Acc}(\tau_i) = \frac{1}{T_i} \sum_{t=0}^{T_i-1} \mathbbm{1}\left(a_t^{(i)} \in \pi^*(s_t^{(i)})\right)
\end{align}

where $\mathbbm{1}(\cdot)$ is the indicator function, $a_t^{(i)}$ is the action taken at step $t$ of trajectory $i$, and $\pi^*(s_t^{(i)})$ is the set of optimal actions for that state.

The grid-level accuracy is then computed by averaging the trajectory-level accuracies across all $N$ generated trajectories:\looseness-1

\begin{align}
    \text{Acc}_{\text{grid}} = \frac{1}{N} \sum_{i=1}^{N} \text{Acc}(\tau_i)
\end{align}

\subsection{Uncertainty Metrics} We also compute several metrics to measure the agent's uncertainty. First, we measure the entropy of the agent's policy $H_{\pi_\theta}$ and the Jensen-Shannon Divergence $\text{JSD}_{\pi_\theta}$ from the optimal policy:

\begin{align}
    H_{\pi_\theta} \coloneq \frac{1}{|\mathcal{S}_{\text{visited}}|} \sum_{s \in \mathcal{S}_{\text{visited}}} H\left(\pi_\theta(\cdot|s)\right)
\end{align}

\begin{align}
    \text{JSD}_{\pi_\theta} \coloneq \frac{1}{|\mathcal{S}_{\text{visited}}|} \sum_{s \in \mathcal{S}_{\text{visited}}} \text{JSD}\left(\pi_\theta(\cdot|s) \parallel \pi^*(s)\right)
\end{align}
with $\mathcal{S}_{\text{visited}}$ being the set of all unique states encountered across all trajectories. We compute the agent's policy $\pi_{\theta}$ empirically by assigning probabilities proportional to action counts during all trajectories for a grid. 

We prefer this method over using the log-probability of the action token because reasoning models like GPT-OSS-20B usually have deliberated and already locked in a final choice during reasoning, thus making the log-probability uninformative in terms of uncertainty.

We also compute the Expected Calibration Error (ECE) over the aggregate counts of all state-action pairs encountered by the agent. Let $\mathcal{D}$ be the collection of all pairs $(s, a)$ from all trajectories for a grid $G$. We partition $\mathcal{D}$ into $M$ disjoint bins $B_1, \dots, B_M$ based on the agent's policy confidence $\pi_\theta(a|s)$.

The ECE is the weighted average of the difference between the average confidence and the empirical accuracy within each bin:

\begin{equation}
    \text{ECE} = \sum_{m=1}^M \frac{|B_m|}{N_{\text{total}}} \left| \text{acc}(B_m) - \text{conf}(B_m) \right|
\end{equation}

where $N_{\text{total}}$ is the total number of state-action pairs. The bin accuracy, representing the empirical probability of optimality, is defined as:

\begin{equation}
    \text{acc}(B_m) = \frac{1}{|B_m|} \sum_{(s, a) \in B_m} \mathbbm{1}\left(a \in \pi^*(s)\right)
\end{equation}

and the bin confidence is the average policy probability:

\begin{equation}
    \text{conf}(B_m) = \frac{1}{|B_m|} \sum_{(s, a) \in B_m} \pi_\theta(a|s)
\end{equation}

In our experiments, we use $M=10$ bins.

Let $\tau_1$ and $\tau_2$ be two different trajectories having the same starting state $s_0$ in grid $G$. We measure the overlap of their unique state-action pairs using the Jaccard Similarity Index:

$$
J(\tau_1, \tau_2) = \frac{\left| S(\tau_1) \cap S(\tau_2) \right|}{\left| S(\tau_1) \cup S(\tau_2) \right|}
$$
where $S(\tau)$ denotes the set of unique states in trajectory $\tau$.

\section{Additional Behavioural Evaluation Results}\label{app:additional_objective_results}
\begin{figure}[H]
    \centering
    \begin{subfigure}[b]{0.45\textwidth}
        \centering
        \includegraphics[width=\linewidth]{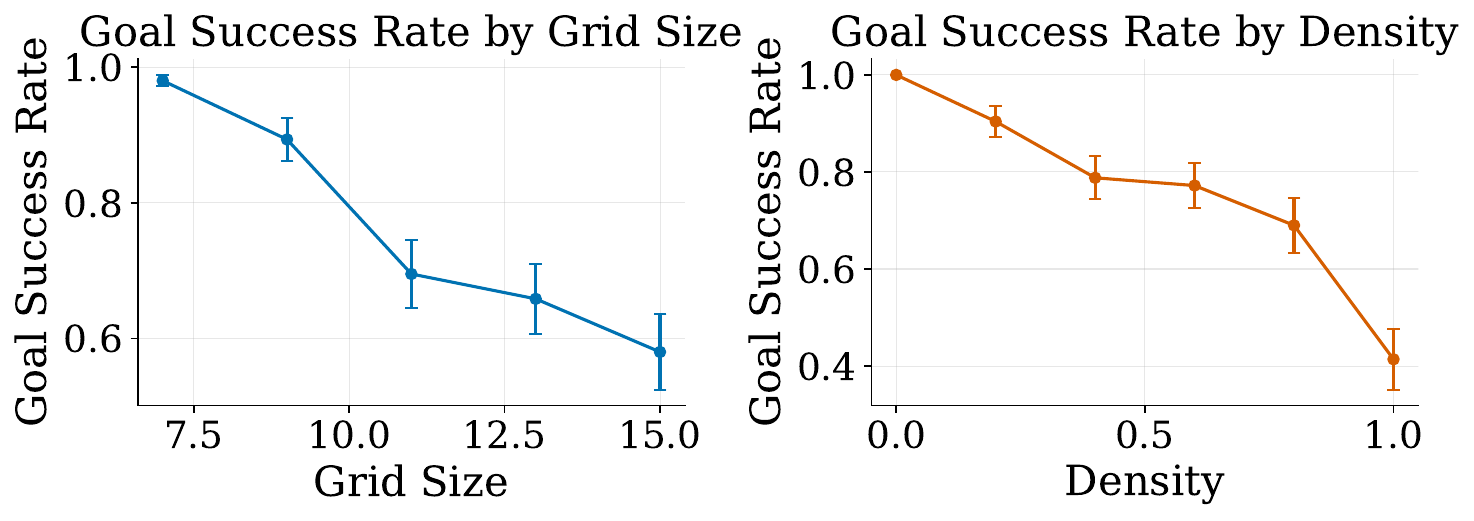}
        \caption{Goal success rate.}
        \label{fig:success_rate}
    \end{subfigure}
    \hfill
    \begin{subfigure}[b]{0.45\textwidth}
        \centering
        \includegraphics[width=\linewidth]{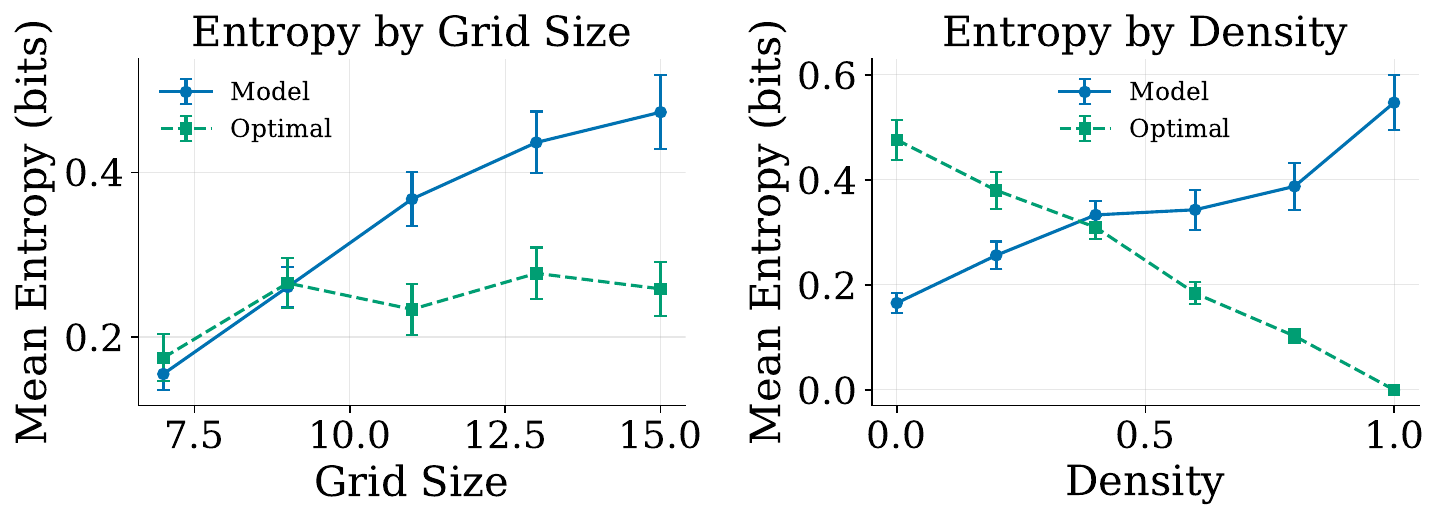}
        \caption{Entropy of the agent's policy.}
        \label{fig:entropy}
    \end{subfigure}

    \begin{subfigure}[b]{0.45\textwidth}
        \centering
        \includegraphics[width=\linewidth]{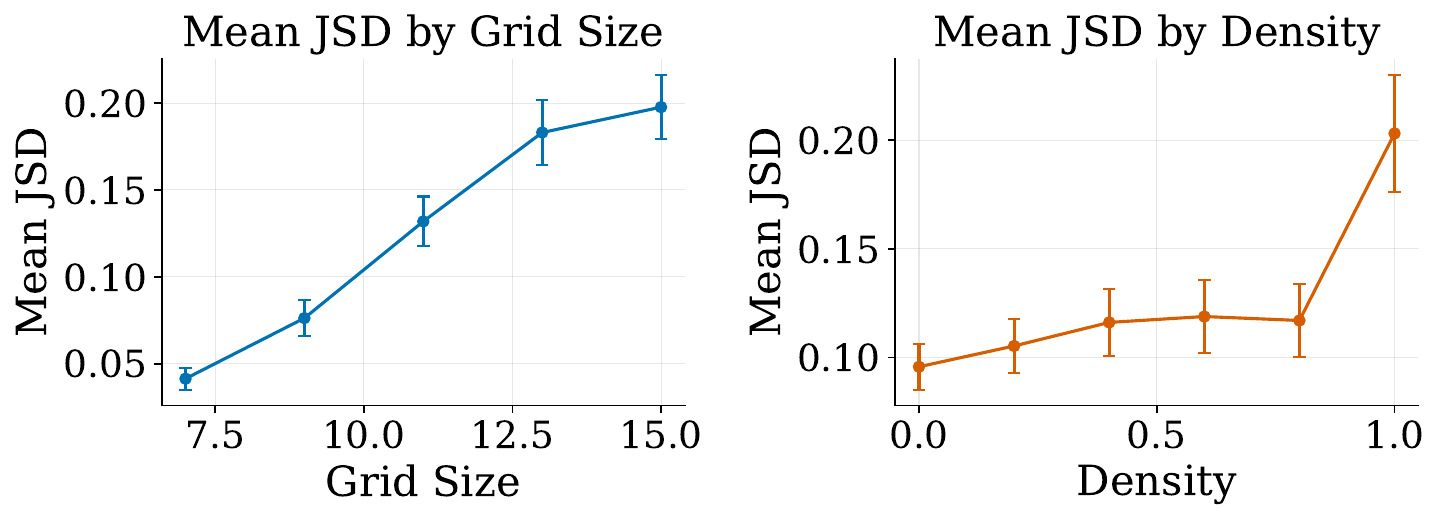}
        \caption{Jensen-Shannon Divergence.}
        \label{fig:jsd}
    \end{subfigure}
    \hfill
    \begin{subfigure}[b]{0.45\textwidth}
        \centering
        \includegraphics[width=\linewidth]{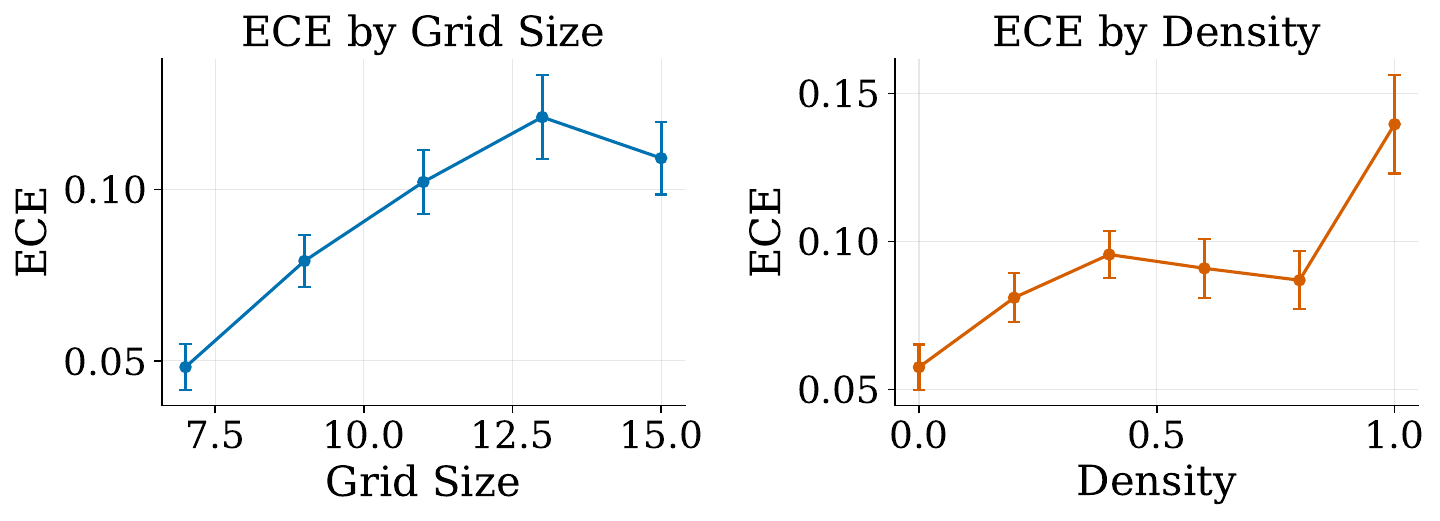}
        \caption{Expected Calibration Error.}
        \label{fig:ece}
    \end{subfigure}
    
    \caption{Performance metrics by size and complexity. (a) Goal success rate, (b) Policy entropy, (c) JSD from optimal policy, and (d) Expected Calibration Error.}
    \label{fig:full_obs_analysis}
    \vspace{-0.8em}
\end{figure}

\section{Iso-difficulty Transform Quantitative Results}\label{app:iso_transform_results}

\begin{figure}[!ht]
    \centering
    \includegraphics[width=0.8\linewidth]{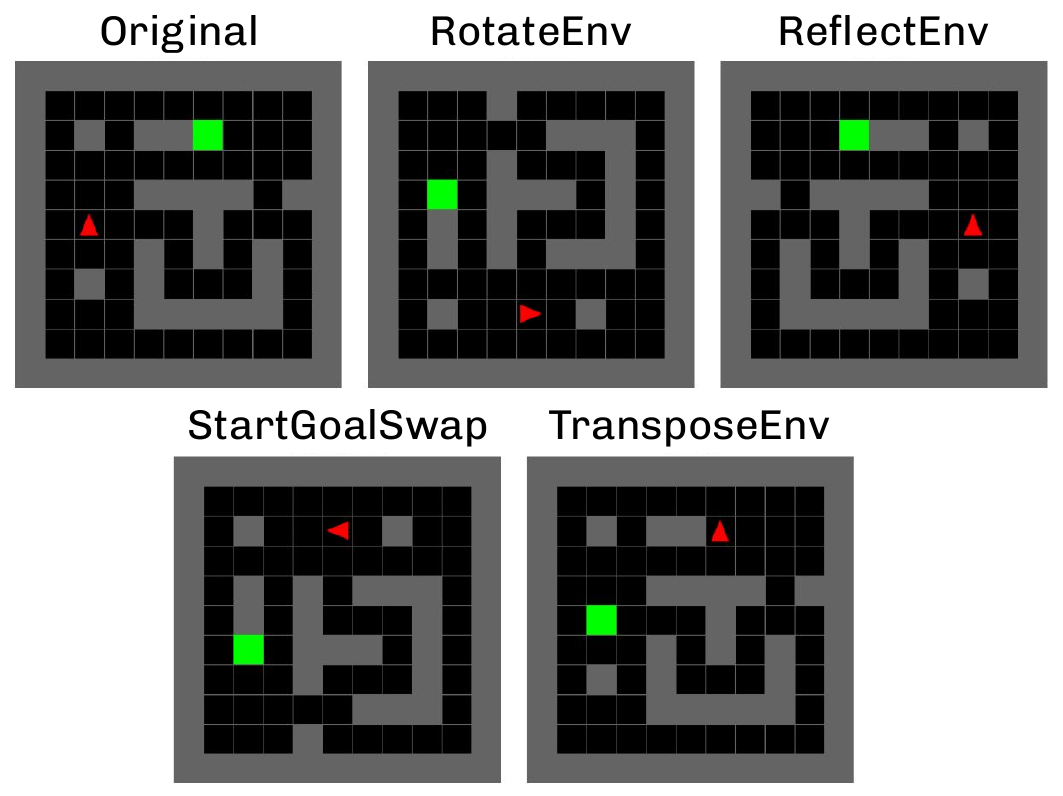}
    \caption{Examples of iso-difficulty transformations.}
    \label{fig:example_transforms}
    \vspace{-0.8em}
\end{figure}

\begin{figure}[!ht]
    \centering
    \includegraphics[width=0.85\linewidth]{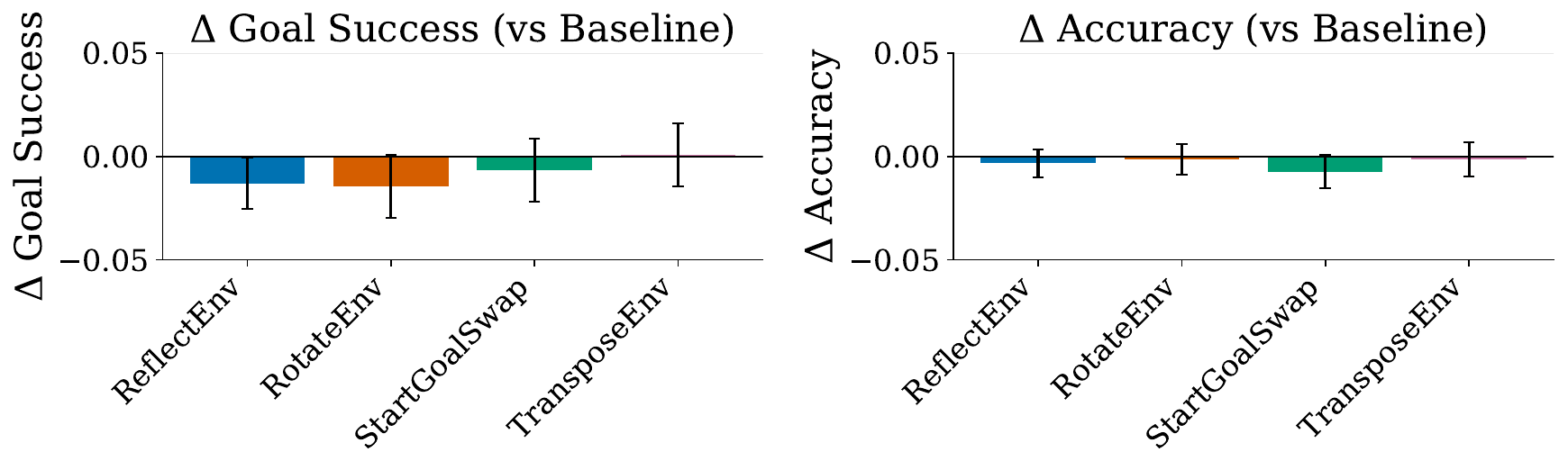}
    \caption{Robustness to Iso-difficulty Environment Transformations.}
    \label{fig:iso_transforms}
    \vspace{-0.8em}
\end{figure}

\begin{table}[H]
\centering
\caption{Statistical significance and effect sizes for iso-difficulty transformations ($N=300$ pairs); Wilcoxon signed-rank test. No significant difference is observed across all transformations.}
\label{tab:performance_stats}
\resizebox{\columnwidth}{!}{
\begin{tabular}{llcc}
\toprule
\textbf{Metric} & \textbf{Transformation} & \textbf{$p$-value} & \textbf{Effect Size} \\
\midrule
\multirow{4}{*}{GSR} 
 & Reflection & 0.582 & 0.059 \\
 & Rotation & 0.311 & 0.103 \\
 & Start/Goal Swap & 0.391 & 0.085 \\
 & Transposition & 0.949 & 0.006 \\
\midrule
\multirow{4}{*}{Accuracy} 
 & Reflection & 0.784 & 0.022 \\
 & Rotation & 0.838 & 0.016 \\
 & Start/Goal Swap & 0.179 & 0.109 \\
 & Transposition & 0.602 & 0.042 \\
\bottomrule
\end{tabular}
}
\vspace{-0.8em}
\end{table}

\section{Additional Representational Evaluation Results}
\label{app:additional-representational-results}

\subsection{Cognitive Map Encoding across Layers}
\label{sec:cogmap-across-layers}

\begin{figure}[t!]
  \centering
  \includegraphics[width=\linewidth]{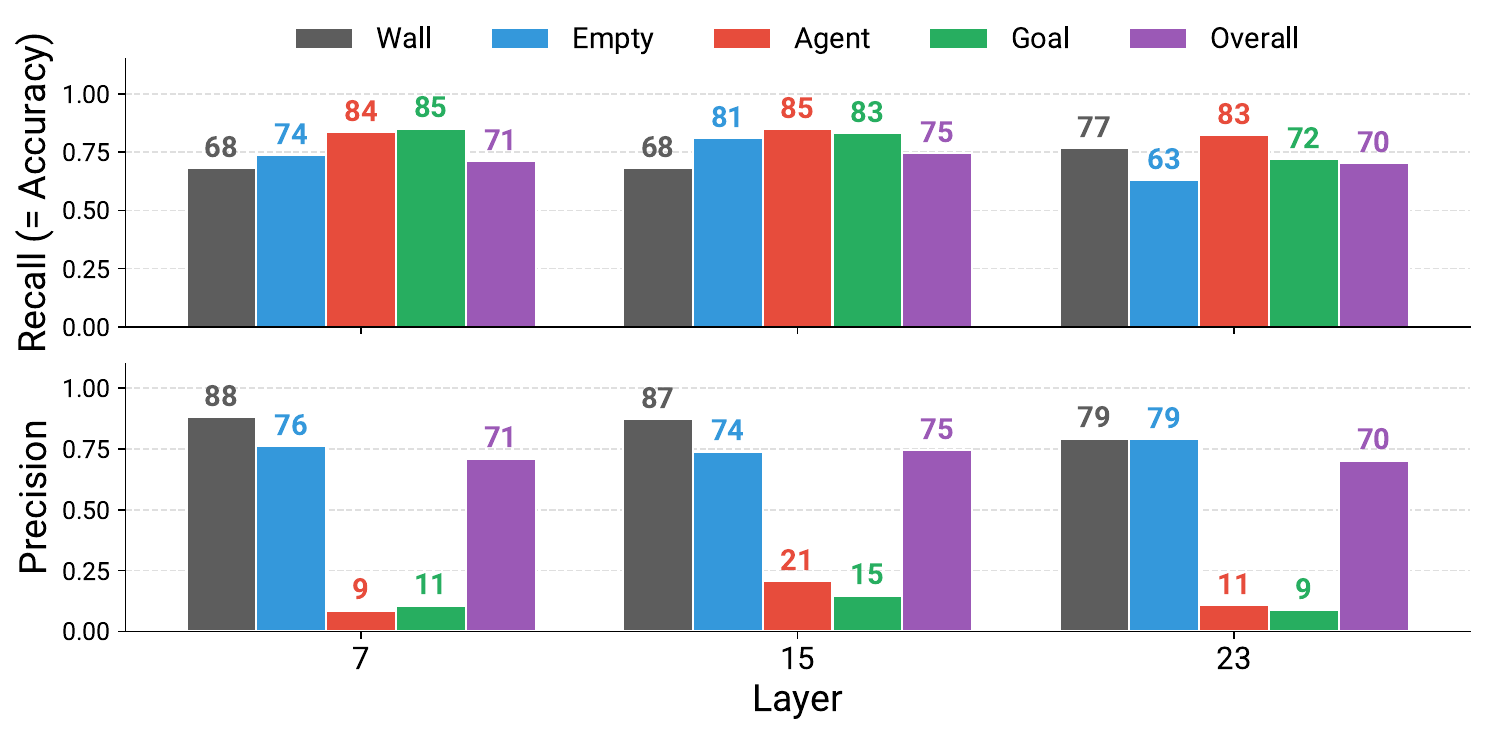}
  \caption{Performance of cognitive map probes across layers. Middle layer activations show the highest overall accuracy, but goal-specific information is already present in early layers.}
  \label{fig:per_layer_metrics}
\end{figure}

We examine how world model information develops across layers in GPT-OSS-20B by training MLP probes on activations from early (7), middle (15), and late (23) layers for grid size 11. Detailed probe performance is reported in \Cref{fig:per_layer_metrics}. Goal-specific spatial information is already present in early layers, but the precision of the \agent{} and \goal{} classes continues to increase in layer 15. By the later layers, overall accuracy declines again, along with recall and precision for \agent{} and \goal{}. This pattern suggests that, at end-of-prompt-token indices, spatial information is most explicitly represented at intermediate layers and is subsequently transformed for the computation of other features rather than being directly preserved for next-token prediction.

\subsection{Size-specific Cognitive Map Probing Results}
\label{sec:cogmap-across-sizes}

\begin{figure}[t!]
  \centering
  \includegraphics[width=\linewidth]{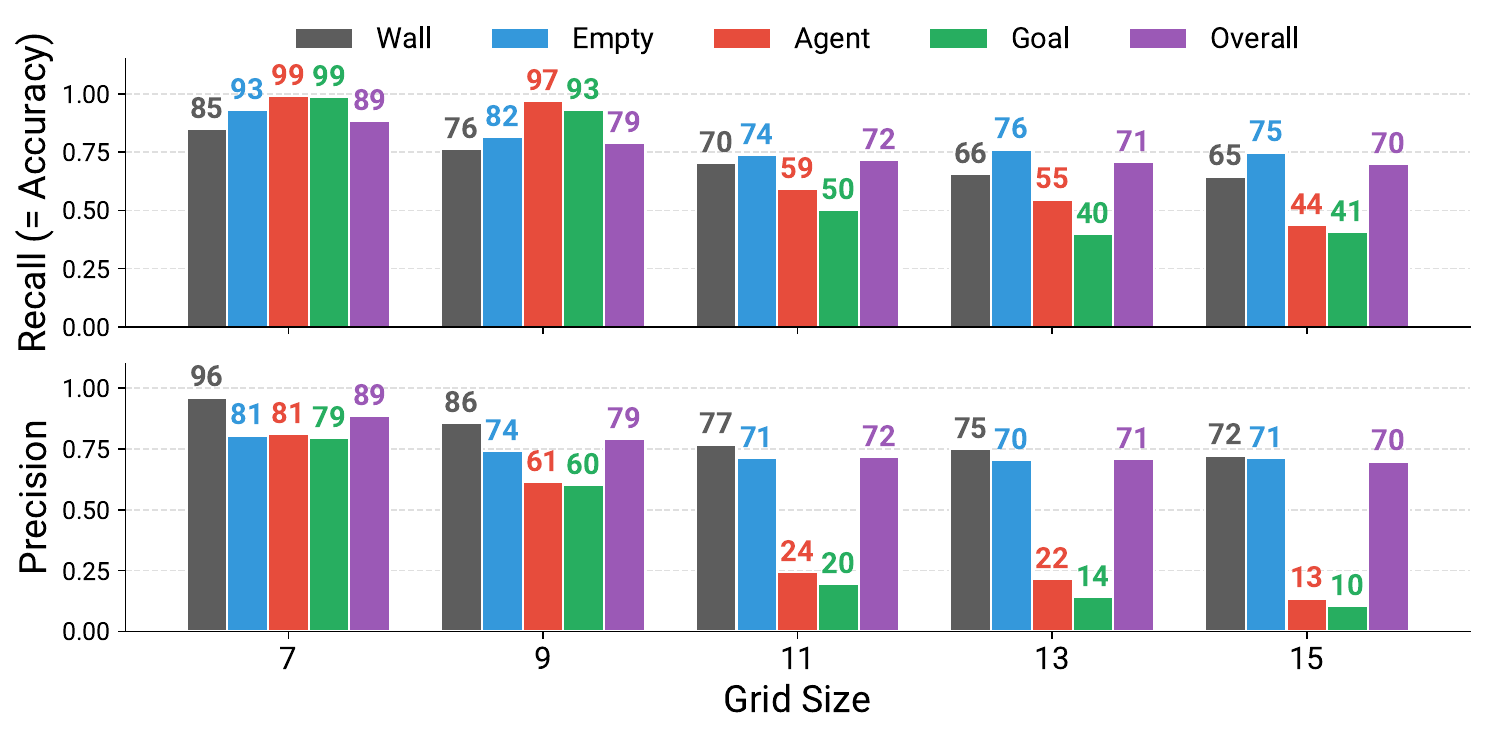}
  \caption{Performance of size-specific MLP probes. Size-specific probes provide comparable performance, but less flexibility than the size-agnostic approach.} 
  \label{fig:size_specific_metrics}
\end{figure}

In the main experiments we trained a single MLP probe on data from all grid sizes, padding inputs to size 15. While this approach has the obvious advantage that we can use the same probe for any grid world, it might result in worse performance than size-specific classifiers.
We test this by training size-specific MLP probes without padding for each grid size. Results are shown in \Cref{fig:size_specific_metrics}. Compared to \Cref{fig:cognitive_maps_overview} (centre), size-specific probes achieve higher overall accuracy for smaller grids, but match the performance of general probes for grid sizes 11–15. Precision for the \agent{} and \goal{} classes is higher across grid sizes, but recall is substantially lower for grid sizes 11–15.
Given these uneven performance differences, together with the flexibility of the size-agnostic approach, we adopt the size-independent MLP probe in our main experiments.

\subsection{Goal Distance Probing Results}
\label{sec:goal-distance-representation}

We also test if internal representations of GPT-OSS-20B encode information about the distance between the agent and the goal.
As with cell identity, distance to the goal is a variable we hypothesise may be tracked because of its direct relevance for goal-directed behaviour. 
Indeed, in our grid worlds, the optimal state value is a monotonic function of the distance to the goal (cf.\ \Cref{sec:fo_gridworld}).
To test this, we train linear and MLP probes on both pre- and post-reasoning activations, 
using the length of the optimal trajectory computed via $A^*$ as the ground-truth distance label. 

\begin{table}[t!]
\small
\centering
\caption{\textbf{Distance Probe Performance}. }
\label{tab:distance_probes_layer15}
\begin{tabular}{llcc}
\toprule
Probe Type & Reasoning Stage & MAE $\downarrow$ & $R^2$ $\uparrow$ \\
\midrule
MLP Probe & Pre-Reasoning & 3.16 & 0.40 \\
 & Post-Reasoning & \textbf{2.67} & 0.36 \\
\midrule
Linear Probe & Pre-Reasoning & 3.40 & 0.42 \\
 & Post-Reasoning & 3.13 & \textbf{0.47} \\
\bottomrule
\end{tabular}
\end{table}

Results are reported in \Cref{tab:distance_probes_layer15}. Across conditions, goal distance can be decoded with a mean absolute error of approximately 3 steps, both before and after reasoning.
The best performance is achieved by the post-reasoning MLP probe, with a mean absolute error of 2.67.
These results suggest that, by reasoning about the grid and planning the next steps, the model develops a more accurate representation of the distance it needs to cover to reach the goal.

\begin{table}[t!]
\centering
\small
\setlength{\tabcolsep}{6pt}
\renewcommand{\arraystretch}{1.1}
\caption{\textbf{Sequence length analysis for one-shot plan decoding.}}
\label{tab:sequence_length_analysis}
\resizebox{0.9\columnwidth}{!}{
\begin{tabular*}{\linewidth}{@{\extracolsep{\fill}} lcc @{}}
\toprule
Metric & Pre-reasoning & Post-reasoning \\
\midrule
Exact length match (\%) $\uparrow$        & 15.71 & 18.85 \\
Predicted length (avg)                   & 4.81  & 5.05 \\
Ground-truth length (avg)                & 4.93  & 4.93 \\
Length bias (avg, pred $-$ true)           & $-0.12$ & $+0.12$ \\
Median abs.\ length error (steps)        & 1.0   & 1.0 \\
\bottomrule
\end{tabular*}}
\end{table}

\subsection{Inverse Recovery Rate}
\label{sec:app-inverse-recovery-rate}
To complement the analysis in \Cref{subsec:policies_against_decoded_beliefs} and \Cref{tab:results_behavior_in_decoded_grid}, we also compute a \textit{reverse recovery} statistic, i.e., the proportion of actions that are suboptimal with respect to the decoded cognitive map but optimal with respect to the ground truth. Reverse recovery declines as grid size and obstacle density increase (from 86.6\% at $7 \times 7$
to 61.1\% at $15 \times 15$, and from 100\% at $d=0.0$ to 60.9\% at $d=1.0$). We interpret this as reflecting growing uncertainty in the agent's internal world state representation as environment complexity increases.

\subsection{Additional Plan Decoder Results}
\label{sec:app-plan-decoder-probe}

We analyse predicted trajectory lengths (\cref{tab:sequence_length_analysis}).
Post-reasoning decoding achieves a higher exact length match rate (19\% vs.\ 16\%) and exhibits a small positive bias in average length (avg pred $-$ true $=+0.12$), whereas pre-reasoning decoding exhibits a small negative bias ($-0.12$). Median absolute length error is 1 step in both settings.

\section{Interventional Analysis via Activation Patching}
\label{app:activation-patching}
To complement the representational analyses in the main paper, we perform activation patching on GPT-OSS-20B using counterfactual trajectories in which either the agent position or the goal position was moved and examine whether the model's predicted action changes. 
We ensure that the optimal action set is disjoint in the counterfactual set and that the agent picks an optimal action in both the original and counterfactual cases, before any patching, to eliminate performance as a confounder.

We intervene at four token-level sites: the full serialised grid state, only the modified grid cells, the pre-reasoning boundary, and the post-reasoning boundary, the last two of which are identical to the probing sites we use. We evaluate patching at three representative layers (7, 15, and 23), as well as all layers at once, in both original $\rightarrow$ counterfactual and counterfactual $\rightarrow$ original directions. For simplicity, we generate counterfactuals for size 7 grids, and filtered for the disjoint constraint described earlier.

The single-layer intervention results are uniformly negative: across 456 executed runs, patching never changes the model's predicted action ($0/456$). This holds for all four intervention sites and for each tested layer individually, and is consistent for both directions. 
In contrast, we find that patching across all layers was always effective in changing the model's predicted action. 
Separately, we also design a control in which the optimal action should not change as a result of the intervention; we find patching here to be perfectly stable: the output action does not change ($144/144$). This indicates that patching does not degrade the model when the source and target already supported the same action. 

Taken together, these results suggest that the information supporting action selection is not causally localised in any single tested layer, even when the relevant grid tokens are directly targeted. This is consistent with the interpretation that task-relevant state information is distributed across layers: it is sufficiently decodable for probing analyses, but not easily steerable through narrow single-layer interventions. Further work is needed to isolate minimal causal interventions in this environment.

\section{Evaluation Settings and Prompts}\label{app:settings}

\subsection{Evaluation Parameters for Behavioural Evaluation}
\Cref{tab:model_config} summarises the main model configuration details.
We use ``low'' reasoning because the GPT-OSS-20B will not finish reasoning on high density / large size grids, even with 10,000 tokens. Note that the reasoning length parameter does not impose a strict token cutoff for GPT-OSS-20B. 

\begin{table}[H]
    \centering
    \caption{\textbf{Model Configuration Details.}}
    \label{tab:model_config}
    \begin{tabular}{ll}
        \toprule
        \textbf{Parameter} & \textbf{Value} \\
        \midrule
        Model ID & \texttt{openai/gpt-oss-20b} \\
        Provider & \texttt{together\_ai} \\
        Interface & \texttt{litellm} \\
        Max Tokens & 10,000 \\
        Temperature & 0.7 \\
        Reasoning Effort & \texttt{low} \\
        Top P & 0.95 \\
        Top Logprobs & 5 \\
        Num. Trajectories per Grid & 10 \\
        \bottomrule
    \end{tabular}
\end{table}

\subsection{Prompt for Behavioural Evaluation}

\begin{lstlisting}[style=prompt]
# Instructions

You are controlling an agent in a grid-based environment with full observability. The agent can move in four directions: up, down, left, and right. The environment contains walls, open spaces, and a goal location. The following symbols are used in the grid representation:

Legend:
---------------
#: Wall
_: Open Space (can be visited)
G: Goal
A: Current agent position
----------------

Your objective is to navigate from the current position (A) to the goal (G) while avoiding walls (#). Importantly, the agent should aim to reach the goal using the least amount of steps possible. You must decide your next move based on the provided information.

You will receive the current state of the grid as a NxM matrix of symbols separated by whitespaces, and with coordinates for each row and column. For example, given the 4x5 grid:

  0 1 2 3 4
0 # # # # #
1 # _ _ G #
2 # A _ _ #
3 # # # # #

The goal is in position (1,3), while the agent's position A is (2,1).

The agent's possible actions at each step are:

Actions:
- UP: Move Up
- DOWN: Move Down
- LEFT: Move Left
- RIGHT: Move Right

Your final answer should be a valid JSON object provided in the following form:

```
{
  "action": "<UP|DOWN|LEFT|RIGHT>",
}
```

DO NOT INCLUDE ANY `json` or `jsonb` in your response NOR TICK MARKS LIKE THIS: ```json ```. Start with { and end with } exactly.

# Inputs

Current grid state:

{{grid_state}}
\end{lstlisting}

\newpage
\subsection{Prompt for Instrumental and Implicit Goals}
\label{sec:prompt_instr_impl}

\begin{lstlisting}[style=prompt]
# Instructions

You are controlling an agent in a grid-based environment with full observability. The agent can move in four directions: up, down, left, and right. The environment contains walls, open spaces, a goal location, and may include doors and keys. The following symbols are used in the grid representation:

Legend:
---------------
#: Wall
_: Open Space (can be visited)
G: Goal
A: Current agent position
D: Door (locked)
K: Key (can unlock doors)
----------------

Your objective is to navigate from the current position (A) to the goal (G) while avoiding walls (#). Importantly, the agent should aim to reach the goal using the least amount of steps possible. You must decide your next move based on the provided information.

## Key and Door Mechanics

- **Key Pickup**: The key is automatically picked up when the agent moves to the same cell as the key (K). Once picked up, the agent carries the key with them.
- **Door Opening**: A locked door (D) is automatically opened and removed from the grid when the agent is adjacent to the door (in front of it) and has already picked up the key. The agent does not need to be on the same cell as the door to open it.
- **Important**: You must pick up the key before you can open the door. If a door blocks your path to the goal, you need to first navigate to the key, pick it up, then navigate to be adjacent to the door to open it.

You will receive the current state of the grid as a NxM matrix of symbols separated by whitespaces, and with coordinates for each row and column. For example, given the 6x7 grid:

  0 1 2 3 4 5 6
0 # # # # # # #
1 # _ _ _ _ _ #
2 # A _ _ _ K #
3 # _ # D # _ #
4 # _ # G # _ #
5 # # # # # # #

The goal is in position (4,3), the agent's position A is (2,1), the door D is at (3,3), and the key K is at (2,5).

The agent's possible actions at each step are:

Actions:
- UP: Move Up
- DOWN: Move Down
- LEFT: Move Left
- RIGHT: Move Right

Your final answer should be a valid JSON object provided in the following form:

```
{
  "action": "<UP|DOWN|LEFT|RIGHT>",
}
```

DO NOT INCLUDE ANY `json` or `jsonb` in your response NOR TICK MARKS LIKE THIS: ```json ```. Start with { and end with } exactly.

# Inputs

Current grid state:

{{grid_state}}

Agent status:
- Carrying key: {{carrying_key}}
\end{lstlisting}
\clearpage

\end{document}